\documentclass[lettersize,journal]{IEEEtran}
\ifCLASSOPTIONcompsoc
  \usepackage[nocompress]{cite}
\else
  \usepackage{cite}
\fi
\usepackage{graphicx}
\usepackage{subcaption}
\usepackage{xspace}
\usepackage{orcidlink}
\usepackage{color}
\usepackage{xcolor}
\usepackage{amsmath}
\usepackage{stmaryrd}
\usepackage{bm}
\usepackage{amssymb}
\usepackage{booktabs}       % professional-quality tables
\usepackage{array,multirow}
\usepackage{tabularx}
\newcolumntype{Y}{>{\centering\arraybackslash}X}

\usepackage{collcell}
\newcolumntype{M}{>{$\displaystyle}c<{$}} % mathematics column
\newcommand\AddLabel[1]{\refstepcounter{equation}(\theequation)\label{#1}}
\newcolumntype{L}{>{\collectcell\AddLabel}r<{\endcollectcell}}% labeled

\usepackage{pifont}
\usepackage{dsfont}
\usepackage{pgfplots}
\usepackage{float}
\usepackage{tikz}
\newcommand*\circled[1]{\tikz[baseline=(char.base)]{
            \node[shape=circle,draw,inner sep=1.2pt] (char) {#1};}}

\usepackage[capitalize]{cleveref}
\crefname{section}{Sec.}{Secs.}
\crefname{equation}{Eq.}{Eqs.}
\crefname{figure}{Fig.}{Figs.}
\crefname{table}{Tab.}{Tabs.}

\usepackage[subtle]{savetrees}
\colorlet{Green}{green!80!black}
\definecolor{amethyst}{rgb}{0.6, 0.4, 0.8}

\def\eg{\textit{e.g.}\xspace}
\def\vs{\textit{v.s.}\xspace}
\def\ie{\textit{i.e.}\xspace}
\def\etc{\textit{etc.}\xspace}

\newcommand{\Labs}{\mathcal{L}_\text{DG}}
\newcommand{\lclust}{\mathcal{L}_\text{DG}}

\newcommand{\Ldecom}{\mathcal{L}_\text{DG}^*}
\newcommand{\LAP}{\mathcal{L}_\text{AP}}
\newcommand{\LROADMAP}{\mathcal{L}_\text{ROADMAP}}
\newcommand{\AP}{\text{AP}}

\newcommand{\LsupAP}{\mathcal{L}_\text{Sup-AP}}
\newcommand{\LsupRk}{\mathcal{L}_\text{Sup-R@k}}
\newcommand{\LsupNDCG}{\mathcal{L}_\text{Sup-NDCG}}

\newcommand{\lhap}{\mathcal{L}_{\hap}}
\newcommand{\lhaps}{\lhap^s}
\newcommand{\hap}{\mathcal{H}\text{-AP}}
\newcommand{\hrank}{\mathcal{H}\text{-rank}^+}
\newcommand{\rank}{\text{rank}}

\DeclareMathOperator{\rel}{rel}

\newcommand{\cmark}{\ding{51}}
\newcommand{\xmark}{\ding{55}}

\usepackage{xr}

\makeatletter
\newcommand*{\addFileDependency}[1]{% argument=file name and extension
\typeout{(#1)}% latexmk will find this if $recorder=0
\@addtofilelist{#1}
\IfFileExists{#1}{}{\typeout{No file #1.}}
}\makeatother

\newcommand*{\myexternaldocument}[1]{%
\externaldocument{#1}%
\addFileDependency{#1.tex}%
\addFileDependency{#1.aux}%
}

\newcommand{\raone}{\scriptsize{R@1}}

\newcommand{\sap}{\scriptsize{AP}}

\newtheorem{property}{Property}

\ifCLASSINFOpdf
\else
\fi
\myexternaldocument{supp}

\begin{document}
\title{Optimization of Rank Losses for Image Retrieval}

\author{Elias Ramzi\orcidlink{0000-0002-0131-2458}, %~\IEEEmembership{Member,~IEEE,}
        Nicolas Audebert\orcidlink{0000-0001-6486-3102}, %~\IEEEmembership{Fellow,~OSA,}
        Clément Rambour\orcidlink{0000-0002-9899-3201}, %~\IEEEmembership{Fellow,~OSA,}
        André Araujo\orcidlink{0000-0002-4214-6185},
        Xavier Bitot\orcidlink{0009-0006-7254-6178}, %,~\IEEEmembership{Life~Fellow,~IEEE}% <-this % stops a space
        and~Nicolas Thome\orcidlink{0000-0003-4871-3045}%~\IEEEmembership{Fellow,~OSA,}
\IEEEcompsocitemizethanks{
\IEEEcompsocthanksitem Elias Ramzi, Nicolas Audebert and Clément Rambour are with the Cnam.
\IEEEcompsocthanksitem André Araujo is with Google Research.
\IEEEcompsocthanksitem Xavier Bitot is with Coexya.
\IEEEcompsocthanksitem Nicolas Thome is with Sorbonne Université.
}% <-this % stops an unwanted space
\thanks{Manuscript received May 26th, 2023}}


\markboth{Submitted to IEEE TRANSACTIONS ON PATTERN ANALYSIS AND MACHINE INTELLIGENCE}%
{Shell \MakeLowercase{\textit{et al.}}: Bare Demo of IEEEtran.cls for Computer Society Journals}



%\vspace{-5em}
\IEEEtitleabstractindextext{%
\begin{abstract}
In image retrieval, standard evaluation metrics rely on score ranking, \eg average precision (AP), recall at k (R@k), normalized discounted cumulative gain (NDCG). In this work we introduce a general framework for robust and decomposable rank losses optimization. It addresses two major challenges for end-to-end training of deep neural networks with rank losses: non-differentiability and non-decomposability. Firstly we propose a general surrogate for ranking operator, SupRank, that is amenable to stochastic gradient descent. It provides an upperbound for rank losses and ensures robust training. Secondly, we use a simple yet effective loss function to reduce the decomposability gap between the averaged batch approximation of ranking losses and their values on the whole training set. We apply our framework to two standard metrics for image retrieval: AP and R@k. Additionally we apply our framework to hierarchical image retrieval. We introduce an extension of AP, the hierarchical average precision $\hap$, and optimize it as well as the NDCG. Finally we create the first hierarchical landmarks retrieval dataset. We use a semi-automatic pipeline to create hierarchical labels, extending the large scale Google Landmarks v2 dataset. The hierarchical dataset is publicly available at \href{https://github.com/cvdfoundation/google-landmark}{\nolinkurl{github.com/cvdfoundation/google-landmark}}. Code will be released at \href{https://github.com/elias-ramzi/SupRank}{\nolinkurl{github.com/elias-ramzi/SupRank}}.
\end{abstract}

}

\maketitle

\IEEEdisplaynontitleabstractindextext

% \fi
\IEEEpeerreviewmaketitle

\section{Introduction}\label{sec:introduction}

Image retrieval (IR) is a major task in computer vision. The goal is to retrieve ``similar'' images to a query in a database. In modern computer vision this is achieved by learning a space of image representation, \ie embeddings, where ``similar'' images are close to each other.

The performances of IR systems are often measured using ranking-based metrics, \eg average precision (AP), recall rate at k (R@k), Normalized Discounted Cumulative Gain (NDCG). These metrics penalize retrieving non-relevant images before other remaining relevant images.

Although these metrics are suited for image retrieval, their use for training deep neural networks is limited. They have two main drawbacks: i) they are not amenable to stochastic gradient descent (SGD) and thus cannot be used directly to train deep neural networks (DNN), ii) they are not decomposable.

There has been a rich literature to provide proxy losses for the task of image retrieval using tuplet losses~\cite{NIPS2002_c3e4035a,hadsell2006dimensionality,DBLP:conf/eccv/RadenovicTC16,DBLP:journals/ijcv/GordoARL17,wu2017sampling,xuan2020hard,NIPS2016_6b180037,DBLP:journals/ijcv/LawTC17,multi_similarity} or cross entropy based losses~\cite{proxynca,norm_softmax,wang2018cosface,deng2019arcface,fewer_is_more,proxynca++}. There also has been extensive work to create rank losses amenable to gradient descent~\cite{Yue:2007,Mcfee10metriclearning,Mohapatra_2018_CVPR,Durand19,blackbox,histogram_loss,He_2018_CVPR,He_2018_DOAP,fastap,revaud2019learning,Engilberge_2019_CVPR,smoothap,patel2022recall}. They create either coarse upper bounds of the target metric or tighter approximations but loosen the upper bound property which affects final performances.

During rank loss training, the loss averaged over batches generally underestimates its value on the whole training dataset, which we refer to as the \textit{decomposability gap}. In image retrieval, attempts to circumvent the problem involve \textit{ad hoc} methods based on hard batch sampling strategies~\cite{Ge_2018_ECCV,Suh_2019_CVPR,wu2017sampling,NIPS2016_6b180037}, storing all training representations/scores~\cite{xbm,blackboxap} or using larger batches~\cite{fastap,revaud2019learning,patel2022recall}, leading to complex models with a large computation or memory overhead.

\begin{figure*}[ht]
    \centering
    \includegraphics[width=\textwidth]{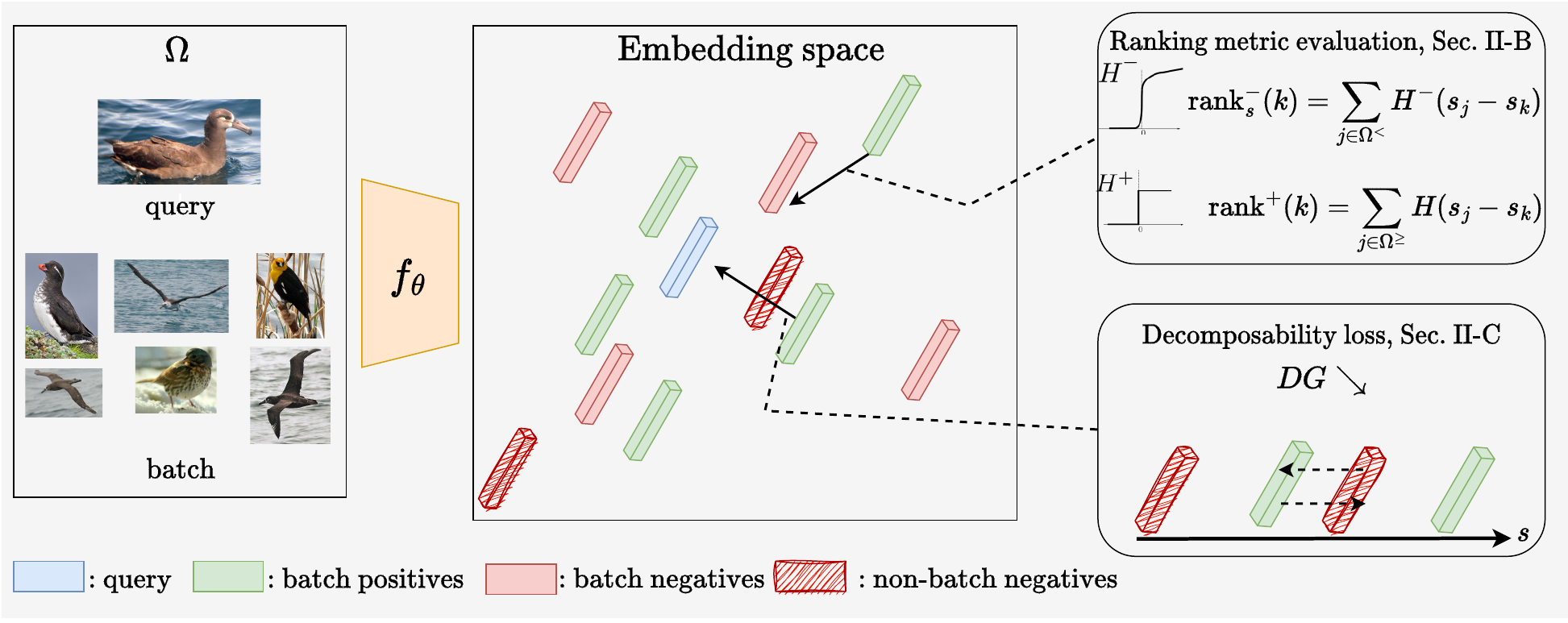}
    \caption{Illustration of our unified framework which supports both hierarchical and non-hierarchical cases. We use a deep neural network $f_\theta$ to embed images. We then optimize its weights in an end-to-end manner using two losses: 1) we optimize the ranking-based evaluation metric using an upper bound approximation of the rank, $\rank^-_s$, as described in~\cref{sec:suprank}, enforcing the batch's positive embeddings to have higher cosine similarity with the query than the batch's negatives; 2) we reduce the decomposability gap, $DG$, of rank losses using a decomposability loss as described in~\cref{sec:decomp}, that supports that positives have higher similarity with the query than all negatives even outside the batch.
    }
    \label{fig:figure_introduction}
    \vspace{-1.3em}
\end{figure*}

The core of our approach is a a unified framework, illustrated in~\cref{fig:figure_introduction} and detailed in~\cref{sec:method}, to optimize rank losses for both hierarchical and standard image retrieval. Specifically, we propose a smooth approximation of the rank which is amenable to SGD and is an upper bound on the true rank, which leads to smooth losses that are upper bounds of the true losses. At training time, we additionally introduce a novel objective to reduce the non-decomposability of smooth rank losses without the need to increase the batch size.

Our framework for end-to-end training of DNN is illustrated in~\cref{fig:figure_introduction}. Using a DNN $f_\theta$ we encode both the query and the rest of the images in the batch. Optimizing the rank loss supports the correct --partial-- ordering in a batch based on our surrogate of the rank, SupRank. Optimizing the decomposability loss supports that the positives will be ranked even before negative items that are not present in the batch. Both losses are amenable to gradient descent, which makes possible to update the model parameters with SGD.

Our framework can be used to optimize rank losses for both hierarchical and non-hierarchical image retrieval. In a first time we show how to instantiate our framework to non-hierarchical image retrieval by optimizing two ranking-based metrics, namely AP and R@k. We show the importance of the two components of our framework in ablation studies. 
Using our AP surrogate, we achieve state-of-the-art image retrieval performances across 3 datasets and 3 neural networks architectures.

In a second instantiation we focus on hierarchical image retrieval~\cite{sun2021dynamic,zheng2022dynamic,ramzi2022hierarchical}. Because metrics used to evaluate fine-grained image retrieval rely on binary labels, \ie similar or dissimilar, they are unable to take into account the severity of the errors. This leads methods that optimize this metrics to lack robustness: they tend to make severe errors when they make errors. Hierarchical image retrieval can be used to mitigate this issue by taking into account non-binary similarity between labels. We introduce the hierarchical average precision, $\hap$, a new metric that extends the AP to non-binary settings. Using our optimization framework, we exhibit how optimizing the $\hap$ and the well known NDCG leads to competitive results for fine-grained image retrieval metrics, while outperforming by large margins both binary methods and hierarchical baselines when considering hierarchical metrics.

Finally we introduce the first hierarchical landmarks retrieval dataset, $\mathcal{H}$-GLDv2, extending the well-known Google Landmarks v2 landmarks retrieval (GLDv2) dataset~\cite{weyand2020google}. While landmarks retrieval has been one of the most popular domain in image retrieval it lacks a hierarchical dataset. $\mathcal{H}$-GLDv2 is a large scale dataset with $1.4$m images and three levels of hierarchies: including $100$k unique landmarks, 78 super-categories and 2 final labels. The labels are publicly available at \href{https://github.com/cvdfoundation/google-landmark}{\nolinkurl{github.com/cvdfoundation/google-landmark}}.

Initial results of our work have been presented in~\cite{ramzi2021robust,ramzi2022hierarchical}. In this work, we unify the methods from these two papers into a framework for the optimization of rank losses, naturally supporting both standard and hierarchical image retrieval problems. Additionally, we include more comprehensive experiments, to consider different decomposability objectives, apply our framework to the recent R@k loss~\cite{patel2022recall} and optimize the NDCG in the hierarchical setting. Finally, in this work we introduce the first hierarchical image retrieval dataset in the domain of landmarks, which is incorporated for a more comprehensive benchmarking of our method.
%\clearpage

\section{Related work}\label{sec:related_work}

\subsection{Image Retrieval proxy losses}
The Image Retrieval community has designed several families of methods to optimize metrics such as AP and R@k. Methods that rely on tuplet-wise losses, like pair losses~\cite{NIPS2002_c3e4035a,hadsell2006dimensionality,DBLP:conf/eccv/RadenovicTC16}, triplet losses~\cite{DBLP:journals/ijcv/GordoARL17,wu2017sampling,xuan2020hard}, or larger tuplets~\cite{NIPS2016_6b180037,DBLP:journals/ijcv/LawTC17,multi_similarity} learn comparison relations between instances. These metric learning methods optimize a very coarse upper bound on AP and need complex post-processing and tricks to be effective. Other methods using proxies have been introduced to lower the computational complexity of tuplet based training~\cite{proxynca,norm_softmax,wang2018cosface,deng2019arcface,fewer_is_more,proxynca++}: they learn jointly a deep model and weight matrix that represent proxies using a cross-entropy based loss. Proxies are approximations of the original data points that should belong to their neighborhood.

\subsection{Rank loss approximations}
Studying smooth rank surrogate losses has a long history. One option for training with rank losses is to design smooth upper bounds.
Seminal works are based on structural SVMs~\cite{Yue:2007,Mcfee10metriclearning}, with extensions to speed-up the "loss-augmented inference"~\cite{Mohapatra_2018_CVPR} or to adapt to weak supervision~\cite{Durand19} were designed to optimize AP. Generic blackbox combinatorial solvers have been introduced~\cite{blackbox} and applied to AP optimization~\cite{blackboxap}. To overcome the brittleness of AP with respect to small score variations, an \textit{ad hoc} perturbation is applied to positive and negative scores  during training. 
These methods provide elegant AP upper bounds, but generally are coarse AP approximations.

Other approaches rely on designing smooth approximations of the the rank function. This is done in soft-binning techniques~\cite{He_2018_CVPR,He_2018_DOAP,histogram_loss,fastap,revaud2019learning} by using a smoothed discretization of similarity scores. Other approaches rely on explicitly approximating the non-differentiable rank functions using neural networks~\cite{Engilberge_2019_CVPR}, or with a sum of sigmoid functions in the Smooth-AP approach~\cite{smoothap} or the more recent Smooth-Recall loss~\cite{patel2022recall}. These approaches enable accurate surrogates by providing tight and smooth approximations of the rank function. However, they do not guarantee that the resulting loss is an upper bound on the true loss. The SupRank introduced in this work is based on a smooth approximation of the rank function leading to an upper bound on the true loss, making our approach both accurate and robust.

\subsection{Decomposability in AP optimization} Batch training is mandatory in deep learning. However, the non-decomposability of AP is a severe issue, since it yields an inconsistent AP gradient estimator. 

Non-decomposability is related to sampling informative constraints in simple AP surrogates, \eg triplet losses, since the constraints' cardinality on the whole training set is prohibitive. This has been addressed by efficient batch sampling~\cite{Harwood_2017_ICCV,Ge_2018_ECCV,Suh_2019_CVPR} or selecting informative constraints within mini-batches ~\cite{NIPS2016_6b180037,VSE++,DBLP:conf/sigir/CarvalhoCPSTC18,Suh_2019_CVPR}. In cross-batch memory technique~\cite{xbm}, the authors assume a slow drift in learned representations to store them and compute global mining in pair-based deep metric learning.

In AP optimization, the non-decomposability has essentially been addressed by a brute force increase of the batch size~\cite{fastap,revaud2019learning,blackbox,patel2022recall}. This includes an important overhead in computation and memory, generally involving a two-step approach for first computing the AP loss and subsequently re-computing activations and back-propagating gradients. In contrast, our loss does not add any overhead and enables good performances for AP optimization even with small batches.

\subsection{Hierarchical predictions and metrics}

There has been a recent regain of interest in Hierarchical Classification (HC)~\cite{dhall2020hierarchical,bertinetto2020making,chang2021your}, to learn robust models that make ``better mistakes" \cite{bertinetto2020making}. However HC is evaluted in  \emph{closed set}, \ie train and test classes are the same. Whereas, hierarchical image retrieval considers the \emph{open set} paradigm, where classes are distinct between train and test sets to better evaluate the generalization abilities of learned models.

The Information Retrieval community uses datasets where documents can be more or less relevant depending on the query~\cite{hjorland2010foundation,graded_relevance}. The quality of their retrieval engine is quantified using ranking based metrics such as the NDCG~\cite{jarvelin2002cumulated,croft2010search}. Several works have investigated how to optimize the NDCG, \eg using pairwise losses~\cite{ranknet} or smooth surrogates~\cite{lambdarank,softrank,Qin2009AGA,bruch2019revisiting}. These works however focused on NDCG, and are without any theoretical guarantees: the surrogates are approximations of the NDCG but not \emph{lower bounds}, \ie their maximization does not imply improved performances during inference. An additional drawback is that NDCG does not relate easily to average precision~\cite{dupret_2011}, the most common metric in image retrieval. Fortunately, there have been some works done to extend AP in a graded setting where relevance between instances is not binary~\cite{robertson2010extending,pap}. The graded Average Precision from~\cite{robertson2010extending} is the closest to our work as it leverages SoftRank for direct optimization of non-binary relevance, although there are significant shortcomings. There is no guarantee that the SoftRank surrogate actually minimizes the graded AP, it requires to annotate datasets with pairwise relevances which is impractical for large scale settings in image retrieval.

Recently, the authors of~\cite{sun2021dynamic} introduced three new hierarchical benchmarks datasets for image retrieval, in addition to a novel hierarchical loss CSL. CSL extends proxy-based triplet losses to the hierarchical setting. However, this method faces the same limitation as triplet losses: minimizing CSL does not explicitly optimize a well-behaved hierarchical evaluation metric, \eg $\hap$. We show experimentally that our method significantly outperforms CSL~\cite{sun2021dynamic} both on hierarchical metrics and AP-level evaluations.

\subsection{Hierarchical datasets}

Hierarchical trees are available for a large number of datasets, such as CUB-200-2011~\cite{CUB}, Cars196~\cite{cars196}, InShop~\cite{liuLQWTcvpr16DeepFashion}, Stanford Online Products~\cite{SOP}, and notably \emph{large-scale} ones such as iNaturalist~\cite{inaturalist}, the three DyML datasets~\cite{sun2021dynamic} and Imagenet~\cite{imagenet}. Hierarchical labels are also less difficult to obtain than fine-grained ones since hierarchical relations can be semi-automatically obtained by grouping fine-grained labels. This was previously done by~\cite{chang2021your} or by using the large lexical database Wordnet~\cite{wordnet} \eg for Imagenet in~\cite{imagenet} and for the SUN database in~\cite{SUN}. In the same spirit, we introduce for the first time a hierarchical dataset for the landmark instance retrieval problem: $\mathcal{h}$-GLDv2. We extend the well-known Google Landmarks Dataset v2~\cite{weyand2020google} with hierarchical labels using a semi-automatic pipeline, leveraging category labels mined from Wikimedia commons and substantial manual cleaning.
%\clearpage

\section{Smooth and decomposable rank losses}\label{sec:method}

\begin{figure*}[t]
    \centering
        
    \begin{subfigure}[t]{0.33\textwidth}
        \centering
        \includegraphics[width=0.75\textwidth]{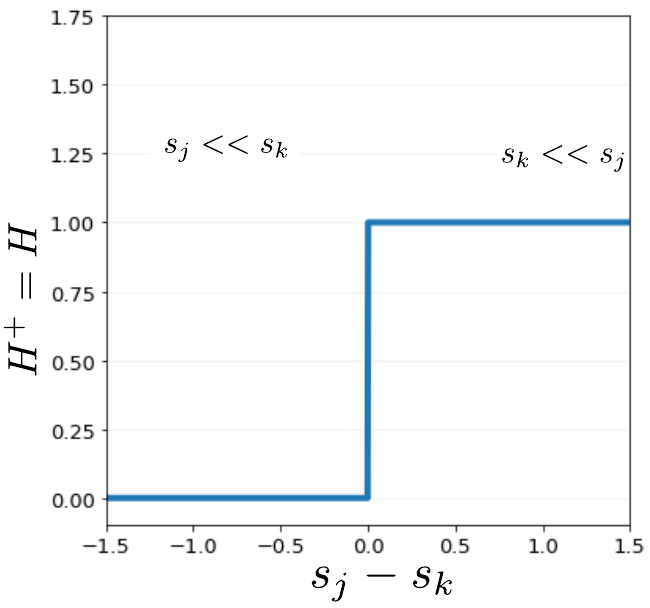}
        \caption{$H^+(x)=H(x)$ in~\cref{eq:definition_rank}}
        \label{fig:h_plus}
    \end{subfigure}%
    \begin{subfigure}[t]{0.33\textwidth}
        \centering
        \includegraphics[width=0.75\textwidth]{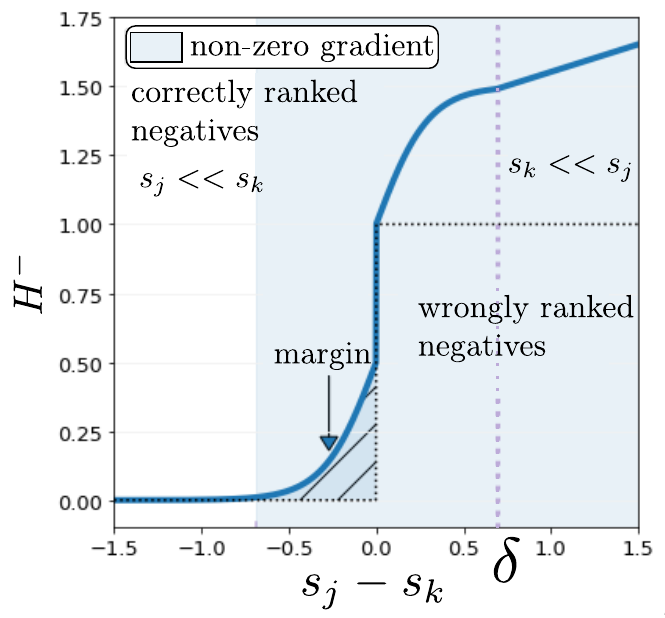}
        \caption{$H^-(x)$ in~\cref{eq:h_minus}}
        \label{fig:h_minus}
    \end{subfigure}%
    \begin{subfigure}[t]{0.33\textwidth}
        \centering
        \includegraphics[width=0.75\textwidth]{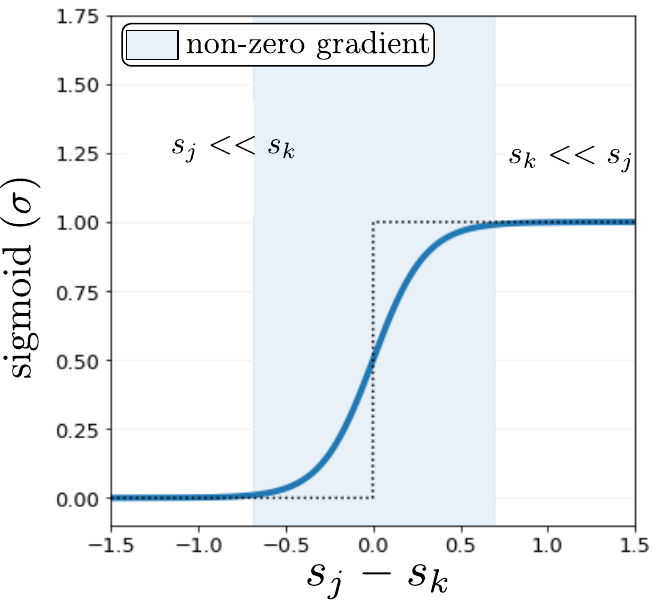}
        \caption{Sigmoid used in~\cite{smoothap}}
        \label{fig:sigmoid}
    \end{subfigure} 
    \vspace{-0.1\intextsep}
    \caption{%Different surrogate losses for the Heaviside (step) function for smooth rank approximation. 
    Proposed surrogate losses for the Heaviside (step): with $H^+(x)$ in \cref{fig:h_plus} and $H^-(x)$ in \cref{fig:h_minus}. Using $H^-$ in~\cref{eq:rank_minus} leads to smooth and upperbounds rank losses. In addition, $H^-(x)$ back-propagates gradients until the correct ranking is satisfied, in contrast to the sigmoid used in \cite{smoothap} (\cref{fig:sigmoid}). }
    \label{fig:smooth_rank}
    \vspace{-1.7em}
\end{figure*}

\subsection{Preliminaries}

Let us consider a retrieval set $\Omega=\left\{\boldsymbol{x_j}\right\}_{j \in \llbracket 1;N\rrbracket}$ composed of $N$ elements, and a set of $M$ queries $\mathcal{Q}$. For each query $\boldsymbol{q_i}$, each element in $\Omega$ is assigned a relevance $\rel(\boldsymbol{x_j},\boldsymbol{q_i}) \in \mathbb{R}$~\cite{hjorland2010foundation}, such that $\rel(\boldsymbol{x_j},\boldsymbol{q_i})>0 $ (resp. $\rel(\boldsymbol{x_j},\boldsymbol{q_i})=0$) if $\boldsymbol{x_j}$ is relevant (resp. irrelevant) with respect to $\boldsymbol{q_i}$. For the standard image retrieval discussed in~\cref{sec:fine_grained_ir}, $\rel(\boldsymbol{x_j},\boldsymbol{q_i}) = 1$ if $x_j$ and $q_i$ share the same fine-grained label and 0 otherwise. In the hierarchical image retrieval setting $\rel(\boldsymbol{x_j},\boldsymbol{q_i})$ models more complex pairwise relevance discussed in~\cref{sec:application_to_hir}. Positive relevance defines the set of positives for a query, \ie $\Omega_i^+ := \left\{\boldsymbol{x_j} \in \Omega | \rel(\boldsymbol{x_j},\boldsymbol{q_i})>0\right\}$. Instances with a relevance of 0 are the negatives, \ie $\Omega_i^- := \left\{\boldsymbol{x_j} \in \Omega | \rel(\boldsymbol{x_j},\boldsymbol{q_i})=0\right\}$.

For each $\boldsymbol{x_j} \in \Omega$, we compute its embedding $\mathbf{v_{\boldsymbol{j}}}\in \mathbb{R}^d$. To do so we use a neural network $f_{\boldsymbol{\theta}}$ parameterized by $\boldsymbol{\theta}$: $\mathbf{v_{\boldsymbol{j}}}:=f_{\boldsymbol{\theta}}(\boldsymbol{x_j})$. In the embedding space $\mathbb{R}^d$, we compute the cosine similarity score between each query $\boldsymbol{q_i}$ and each element in $\Omega$: $s(\boldsymbol{q_i},\boldsymbol{x_j}) = \mathbf{v_{\boldsymbol{q_i}}}^T \mathbf{v_{j}} / ||\mathbf{v_{q_i}}|| \cdot ||\mathbf{v_{j}}||$.

During training, our goal is to optimize, for each query $\boldsymbol{q_i}$, the model parameters $\boldsymbol{\theta}$ such that the ranking, \ie decreasing order of cosine similarity, matches the ground truth ranking, \ie decreasing order of relevances. More precisely, we optimize a ranking-based metric $0\leq\mathcal{M}_i\leq1$ %\andre{Consider writing $\mathcal{M}_i$ here, it would ease readability for me}
that penalizes inversion between positive instances and negative ones. The target loss is averaged over all queries:
\begin{equation}
  \mathcal{L}_{\mathcal{M}}(\boldsymbol{\theta}) = 1-\frac{1}{M} \sum_{i=1}^M  \mathcal{M}_i(\boldsymbol{\theta})
    \label{eq:optimizing_M}
\end{equation}

As previously mentioned, there are two main challenges with SGD optimization of rank losses: i) they are not differentiable with respect to $\boldsymbol{\theta}$, and ii) they do not linearly decompose into batches. We propose to address both issues: we introduce a robust differentiable ranking surrogate, SupRank (\cref{sec:suprank}), and add a decomposable objective (\cref{sec:decomp}) to improve rank losses' behavior in a batch setting. Our final \textbf{RO}bust and \textbf{D}ecomposable (ROD) loss $\mathcal{L}_{\text{ROD-}\mathcal{M}}$ combines a differentiable surrogate loss of a target ranking-based metric, $\mathcal{L}_{\text{Sup-}\mathcal{M}}$, and the decomposable objective $\Labs$ with a linear combination, weighted by the hyper-parameter $\lambda$:

\begin{equation}
\mathcal{L}_{\text{ROD-}\mathcal{M}}(\boldsymbol{\theta}) = (1-\lambda) \cdot \mathcal{L}_{\text{Sup-}\mathcal{M}}(\boldsymbol{\theta}) + \lambda \cdot \Ldecom(\boldsymbol{\theta})
    \label{eq:rod_framework}
\end{equation}

\subsection{SupRank: smooth approximation of the rank}\label{sec:suprank}

The non-differentiablity in rank losses comes from the ranking operator, which can be viewed as counting the number of instances that have a similarity score greater than the considered instance\footnote{For the sake of readability we drop in the following the dependence on $\boldsymbol{\theta}$ for the rank, \ie $\rank(k):=\rank(k,\theta)$ and on the query for the similarity, \ie $s_j:=s(q_i, x_j)$.}, \ie:
\begin{align}
    & \rank(k) = \underbrace{1 + \sum_{j\in \Omega^\geq_{i,k}} H(s_j - s_k)}_{\rank^+(k)} + \underbrace{\sum_{j\in  \Omega_{i,k}^<} H(s_j - s_k)}_{\rank^-(k)}
    \label{eq:definition_rank}
\end{align}

\noindent where $H$ is the Heaviside (step) function $H(t) = 1 \; \text{if} \; t \geq 0,\; 0 \; \text{otherwise}$; $\Omega^\geq_{i,k} = \left\{\boldsymbol{x_p} \in \Omega | \rel(\boldsymbol{x_p},\boldsymbol{q_i})\geq\rel(\boldsymbol{x_k},\boldsymbol{q_i})\right\}$, \ie the set of instances with a relevance greater or equal to $k$'s, and $\Omega^<_{i,k} = \left\{\boldsymbol{x_p} \in \Omega | \rel(\boldsymbol{x_p},\boldsymbol{q_i})<\rel(\boldsymbol{x_k},\boldsymbol{q_i})\right\}$ the set of instances with a relevance strictly lower to $k$'s (in standard IR $\Omega^<_{i,k} = \Omega^-_i$.). Note that for both $\rank^+(k)$ and $\rank^-(k)$ in \cref{eq:definition_rank} $k$ is always positive, \ie in $\Omega^+$, and $x_j$ can either be negative, \ie in $\Omega^-$, in $\rank^-$ or positive in $\rank^+$, \ie in $\Omega^+$. 

From \cref{eq:definition_rank} it becomes clear that the rank is non-amenable to gradient descent optimization due to the Heaviside (step) function $H$ (see~\cref{fig:h_plus}), whose derivatives are either zero or undefined.

\noindent\textbf{SupRank} To provide rank losses amenable to SGD, we introduce a smooth approximation of the rank function. We propose a different behavior between $\rank^+(k)$ and $\rank^-(k)$ in \cref{eq:definition_rank} by defining two functions $H^+$ and $H^-$. For $\rank^+(k)$, we keep the Heaviside function, \ie $H^+=H$ (see Fig. \ref{fig:h_plus}). This ignores $\rank^+(k)$ in gradient-based ranking optimization. It has been observed in other works that optimizing $\rank^-$ is sufficient~\cite{li2022rethinking}. For $\rank^-(k)$ we want smooth surrogate $H^-$ for $H$ that is a amenable to SGD and an upper bound on the Heaviside function. We define the following $H^-$ function, illustrated in Fig~\ref{fig:h_minus}, that is both:
\begin{equation}
    H^-(t) = 
    \begin{cases}
      \sigma(\frac{t}{\tau}) \quad \text{if} \; t \leq 0 \\
      \sigma(\frac{t}{\tau}) + 0.5 \quad \text{if} \; t \in [0;\delta] \quad \text{with} \; \delta \geq 0\\
      \rho \cdot (t - \delta) + \sigma(\frac{\delta}{\tau}) + 0.5 \quad \text{if} \; t > \delta \\
    \end{cases}
    \label{eq:h_minus}
\end{equation}

\noindent where $\sigma$ is the sigmoid function (Fig. \ref{fig:sigmoid}), $\delta$, $\tau$ and $\rho$ are hyper-parameters. $\delta$ is chosen such that the sigmoidal part of $H^-$ reaches the saturation regime and is fixed for the rest of the paper (see supplementary~\cref{sec:choice_of_delta}). We keep $\tau$ as in~\cite{smoothap} and study the robustness to $\rho$ in~\cref{sec:roadmap_hyp}.

From $H^-$ in~\cref{eq:h_minus}, we define the following rank surrogate that can be used plug-and-play for rank losses optimization:
\begin{equation}
    \label{eq:rank_minus}
    \rank_s^-(k)=\sum\limits_{j\in  \Omega^<_{i,k}} H^-(s_j - s_k)
\end{equation}

\textbf{SupRank has two main features:}\vspace{0.1cm}

\noindent$~~~\blacktriangleright$ \textbf{\circled{1} Surrogate losses based on SupRank are upper bound of the target metrics}
, since $H^-$ in \cref{eq:h_minus} is an upper bound of a step function (Fig~\ref{fig:h_minus}). This is an important property, since it ensures that the model keeps training until the correct ranking is obtained. It is worth noting that existing smooth rank approximations in the literature~\cite{histogram_loss,fastap,revaud2019learning,smoothap} do not fulfill this property.

\noindent$~~~\blacktriangleright$ \textbf{\circled{2} SupRank brings training gradients until the correct ranking plus a margin is fulfilled.}
When the ranking is incorrect, an instance with a lower relevance $\boldsymbol{x_j}$ is ranked before an instance of higher relevance $\boldsymbol{x_k}$, thus $s_j > s_k$ and $H^-(s_j-s_k)$ in \cref{eq:h_minus} has a non-zero derivative. We use a sigmoid to have a large gradient when $s_j-s_k$ is small. To overcome vanishing gradients of the sigmoid for large values $s_j - s_k$, we use a linear function ensuring constant $\rho$ derivative. When the ranking is correct ($s_j < s_k$), we enforce robustness by imposing a margin parameterized by $\tau$ (sigmoid in \cref{eq:h_minus}). This margin overcomes the brittleness of rank losses, which vanish as soon as the ranking is correct~\cite{He_2018_CVPR,fastap,blackbox}.

\subsection{Decomposable rank losses}\label{sec:decomp}

As illustrated in \cref{eq:optimizing_M}, rank losses decompose linearly between queries $\boldsymbol{q_i}$, but do not between retrieved instances. We therefore focus our analysis of the non-decomposability on a single query. For a retrieval set $\Omega$ of $N$ elements, we consider $\{\mathcal{B}_b\}_{b\in\{1:K\}}$ batches of size B, such that $N/B=K \in \mathbb{N}$. Let $\mathcal{M}_b(\boldsymbol{\theta})$ be the metric $\mathcal{M}$ in batch $b$ for a query, we define the ``decomposability gap'' $DG$ as: 
\begin{equation}
    \label{eq:decomposability-gap}
    DG(\boldsymbol{\theta}) =  \frac{1}{K}  \sum_{b=1}^K \mathcal{M}_b(\boldsymbol{\theta}) - \mathcal{M}(\boldsymbol{\theta})
\end{equation}
$DG $ in \cref{eq:decomposability-gap} is a direct measure of the non-decomposability of any metric $\mathcal{M}$ \textcolor{black}{(illustrated for AP in~\cref{sec:sup_dg} )}. Our motivation here is to decrease $DG$, \ie to have the average metric over the batches as close as possible to the metric computed over the whole training set. To this end, we use a additional objective during training that aims at reducing the non-decomposability.

\noindent\textbf{Pair-based decomposability loss} We use the following decomposability loss $\Labs$ that was first introduced in ROADMAP~\cite{ramzi2021robust}, and used in other work~\cite{liao2022supervised} to reduce the non-decomposability of ranking losses:

\begin{equation}
    \label{eq:labs}
    \begin{aligned}
    \Labs(\boldsymbol{\theta}) & = \frac{1}{|\Omega^+|}
     \sum_{\boldsymbol{x_j} \in \Omega^+} [\alpha - s_j]_+ + \frac{1}{|\Omega^-|} \sum_{\boldsymbol{x_j} \in \Omega^-} [s_j - \beta]_+
    \end{aligned}
\end{equation}
where $[x]_+ = \max(0,x)$. $\Labs$ is a pair-based loss~\cite{hadsell2006dimensionality}, which we revisit in our context to ``calibrate" the scores between mini-batches. Intuitively, the fact that the positive (resp. negative) scores are above (resp. below) a threshold $\alpha$ (resp. $\beta$) in the mini-batches makes $\mathcal{M}_b$ closer to $\mathcal{M}$, which we support with an analysis in~\cref{sec:sup_upperbound}.

\medbreak
\noindent\textbf{Proxy-based decomposability loss} In HAPPIER~\cite{ramzi2022hierarchical} we used the following proxy-based loss as the decomposability objective:

\begin{equation}\label{eq:cluster_loss}
    \Ldecom(\theta) = - \log\left( \frac{\exp(\frac{v_y^T p_y}{\eta})}{\sum_{{p_z}\in\mathcal{Z}} \exp(\frac{v_y^T p_z}{\eta})} \right),
\end{equation} 
\noindent where $p_y$ is the normalized proxy corresponding to the fine-grained class of the embedding $v_y$, $\mathcal{Z}$ is the set of proxies, and $\eta$ is a temperature scaling parameter. $\Ldecom$ is a classification-based proxy loss~\cite{norm_softmax} that imposes a margin instances and the proxies. $\Ldecom$ has thus a similar effect to $\Labs$ on the decomposability of rank losses. In our experiments we show that both decomposability losses improve ranking losses optimization.
% \clearpage

\section{Instantiation to standard image retrieval}\label{sec:fine_grained_ir}

In this section we apply the framework described previously to standard image retrieval where $\rel(x, q) \in \{0,1\}$. Specifically we show how to directly optimize two metrics that are widely used in the image retrieval community, \ie AP and R@k. 

\subsection{Application to Average Precision}\label{sec:roadmap}

The average precision measures the quality of a ranking by penalizing inversion between positives and negatives. It strongly penalizes inversion at the top of the ranking. It is defined for each query $q_i$ as follows:
\begin{align}
    \label{eq:ap_definition}
    &\text{AP}_i = \frac{1}{|\Omega_i^+|} \sum_{k\in\Omega_i^+} \frac{\rank^+(k)}{\rank(k)}
\end{align}

The overall AP loss $\LAP$ is averaged over all queries: 
\begin{align} \label{eq:average_precision_with_ranks}
  & \LAP(\boldsymbol{\theta}) = 1-\frac{1}{M} \sum_{i=1}^M  \AP_i(\boldsymbol{\theta})
\end{align}

Using our surrogate of the rank, SupRank, we define the following AP surrogate loss:
\begin{equation}
\LsupAP(\boldsymbol{\theta}) = 1-\frac{1}{M} \sum_{i=1}^M \frac{1}{|\Omega^+_{i}|} \sum_{k\in \Omega^+_{i}} \frac{\rank^+(k)}{\rank^+(k)+\rank_s^-(k)}
  \label{eq:SupAP}
\end{equation}

Finally we equip the AP surrogate loss with the $\Labs$ loss to support the decomposability of the AP, yielding our \textbf{RO}bust \textbf{A}nd \textbf{D}eco\textbf{M}posable \textbf{A}verage \textbf{P}recision:
\begin{equation}
\LROADMAP(\boldsymbol{\theta}) = (1-\lambda) \cdot \LsupAP(\boldsymbol{\theta}) + \lambda \cdot \Labs(\boldsymbol{\theta})
    \label{eq:roadmap}
\end{equation}

\subsection{Application to the Recall at k}\label{sec:recall_loss}

Another metric often used in image retrieval is the recall rate at k. In the image retrieval community it is often defined as:

\begin{equation}
    \text{R@k} = \frac{1}{M} \sum_{i=1}^M\mathds{1}(\text{positive element in top-$k$})
\end{equation}

However in the literature the recall is most often defined as:
\begin{equation}
    \text{TR@k} = \frac{1}{M} \sum_{i=1}^M\frac{\#\text{ positive elements in top-$k$}}{\min(k, \#\text{ positive elements})}
\end{equation}

It was shown in~\cite{patel2022recall} that the TR@k can be written similarly to other ranking-based metrics, \ie using the rank, for each query $q_i$ as:
\begin{equation}\label{eq:def_recall}
    \text{TR@k} = \frac{1}{M} \sum_{i=1}^M \frac{1}{\min(|\Omega^+_i|, k)} \sum_{p\in\Omega^+_i} H(k - \rank(p))
\end{equation}

Using the expression of~\cref{eq:def_recall} and SupRank we can derive a surrogate loss function for the recall for a single %\andre{single?}
query as:
\begin{equation}\label{eq:def_recall_loss}
    \LsupRk = 1 - \frac{1}{\min(|\Omega^+|, k)} \sum_{p\in\Omega^+} \sigma(\frac{k - (\rank^+(p)+\rank^-_s(p))}{\tau^*})
\end{equation}
The authors of~\cite{patel2022recall} use different level of recalls in their loss, which we follow \ie $\mathcal{L}_{\text{Sup-R@}\mathcal{K}} = \frac{1}{|\mathcal{K}|} \sum_{k\in \mathcal{K}} \LsupRk$, it is necessary to provide enough gradient signal to all positive items. To train $\mathcal{L}_{\text{Sup-}R@k}$, it is also necessary to approximate a second time the Heaviside function, using a sigmoid with temperature factor $\tau^*$.
We combine it with $\Labs$ yielding the resulting differentiable and decomposable R@k loss:

\begin{equation}\label{eq:def_rod_rk}
    \mathcal{L}_{\text{ROD-R@}\mathcal{K}} = (1-\lambda) \cdot \mathcal{L}_{\text{Sup-R@}\mathcal{K}} + \lambda \cdot \Labs
\end{equation}
% \clearpage

\section{Instantiation to Hierarchical Image Retrieval}\label{sec:application_to_hir}

Standard metrics (\eg AP or R@k) are only defined for binary labels, \ie \emph{fine-grained} labels: an image is negative if it is not strictly similar to the query. These metrics are by design unable to take into account the severity of the mistakes. To mitigate this issue we propose to optimize a new ranking-based metric, $\hap$ introduced in~\cref{sec:hap}, that extends AP beyond binary labels, and the standard NDCG in~\cref{sec:ndcg}.

\begin{figure}[ht]
    \centering
    \includegraphics[width=0.4\textwidth]{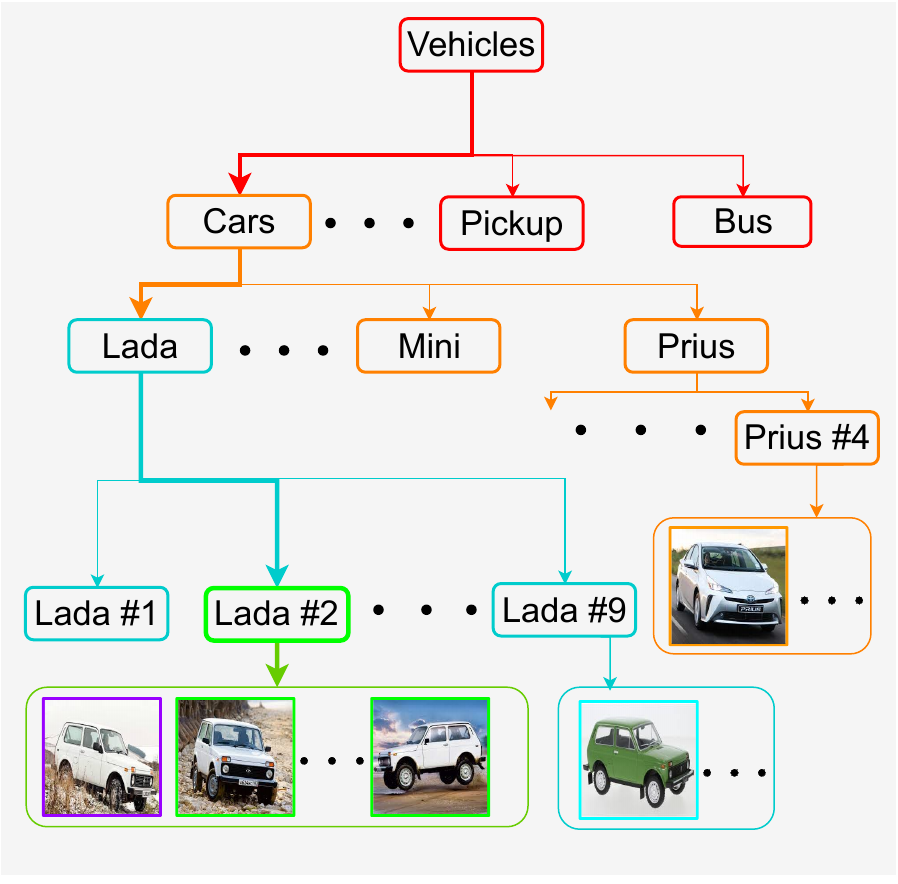}
    \vspace{-0.4\intextsep}
    \caption{We leverage a hierarchical tree representing the semantic similarities between concepts to produce more robust ranking.
    }
    \label{fig:hierarchical_tree}
    \vspace{-1.5em}
\end{figure}

\medbreak
\noindent\textbf{Additional training context}
We assume that we have access to a hierarchical tree defining semantic similarities between concepts as in~\cref{fig:hierarchical_tree}. For a query $\boldsymbol{q}$, we partition the set of retrieved instances into $L+1$ disjoint subsets $\left\{\Omega^{(l)}\right\}_{l \in \llbracket 0;L\rrbracket}$. $\Omega^{(L)}$ is the subset of the most similar instances to the query (\ie fine-grained level): for $L=3$ and a ``Lada \#2'' query (purple), $\Omega^{(3)}$ are the images of the same ``Lada \#2'' (green) in~\cref{fig:hierarchical_tree}. The set $\Omega^{(l)}$ for $l<L$ contains instances with smaller relevance with respect to the query: $\Omega^{(2)}$ in \cref{fig:hierarchical_tree} is the set of ``Lada'' that are not ``Lada \#2'' (blue) and $\Omega^{(1)}$ is the set of ``Cars'' that are not ``Lada'' (orange). We also define $\Omega^- := \Omega^{(0)}$ as the set of negative instances, \ie the set of vehicles that are not ``Cars'' (in red) in~\cref{fig:hierarchical_tree} and $\Omega^+ = \bigcup_{l=1}^L \Omega^{(l)}$. Given a query $q$, we use this partition to define the relevance of $k\in\Omega^{(l)}$, $\rel(k):=\rel(x_k, q)$.

\vspace{-1em}
\subsection{Hierarchical Average Precision}\label{sec:hap}

We propose an extension of AP that leverages non-binary labels. To do so, we extend $\rank^+$ to the hierarchical case with a hierarchical $\rank^+$, $\hrank$:

\begin{equation}
    \hrank(k) = \rel(k) + \sum_{j\in\Omega^+} \min(\rel(k), \rel(j))\cdot H(s_j-s_k) ~.
    \label{eq:hierarchical_rank}
\end{equation}

\begin{figure}[t]
    \centering
    \includegraphics[width=0.5\textwidth]{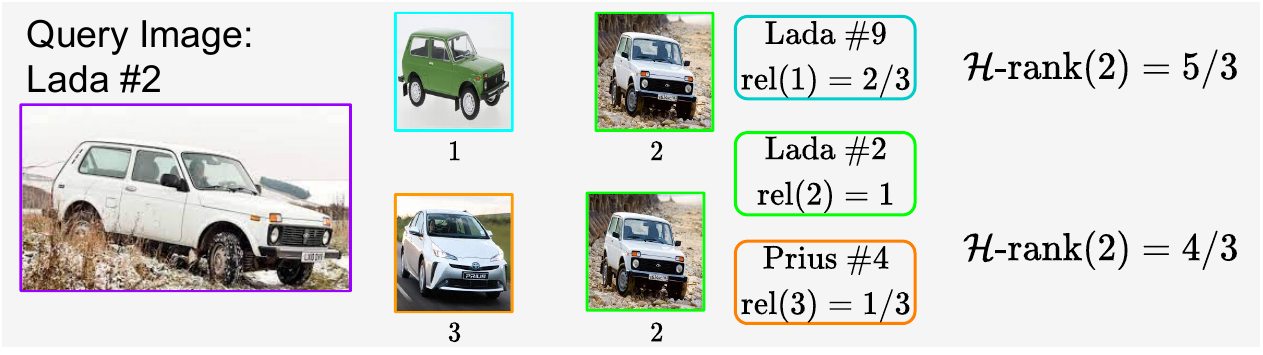}
    \vspace{-0.4\intextsep}
    \caption{
    Given a ``\textcolor{Green}{Lada \#2}'' \textcolor{amethyst}{query}, the top inversion is less severe than the bottom one. Indeed on the top row instance $1$ is semantically closer to the query -- it is a ``\textcolor{cyan}{Lada}''-- than instance $3$ on the bottom row. As instance $3$'s closest common ancestor with the query, ``\textcolor{orange}{Cars}'', is farther in the hierarchical tree~\cref{fig:hierarchical_tree}. This is why $\hrank(2)$ is greater on the top row ($5/3$) than on the bottom row ($4/3$).
    }
    \label{fig:figure_hrank}
    \vspace{-1em}
\end{figure}

Intuitively, $\min(\rel(k), \rel(j))$ corresponds to seeking the closest ancestor shared by instance $k$ and $j$ with the query in the hierarchical tree. As illustrated in~\cref{fig:figure_hrank}, $\hrank$ induces a smoother penalization for instances that do not share the same fine-grained label as the query but still share some coarser semantics, which is not the case for $\rank^+$.

From $\hrank$ in~\cref{eq:hierarchical_rank} we define the Hierarchical Average Precision, $\hap$:
\begin{equation}
\label{eq:def_hap}
    \hap = \frac{1}{\sum_{k\in\Omega^+}\rel(k)} \sum_{k\in\Omega^+} \frac{\hrank(k)}{\rank(k)}
\end{equation}
\cref{eq:def_hap} extends the AP to non-binary labels. We replace $\rank^+$ by our hierarchical rank $\hrank$ and the term $|\Omega^+|$ is replaced by $\sum_{k\in\Omega^+}\rel(k)$ for proper normalization (both representing the ``sum of positives'', see more details in~\cref{sec:sup_normalization_constant_hap}).

$\hap$ extends the desirable properties of the AP. It evaluates the quality of a ranking by: i) penalizing inversions of instances that are not ranked in decreasing order of relevances with respect to the query, ii) giving stronger emphasis to inversions that occur at the top of the ranking. Finally, we can observe that, by this definition, $\hap$ is equal to the AP in the binary setting ($L=1$). This makes $\hap$ a \emph{consistent generalization} of AP (\textcolor{black}{details in~\cref{sec:sup_hap_consistent_ap}}).

\subsubsection{Relevance function design}\label{seq:relevance}

The relevance $\rel(k)$ defines how ``similar'' an instance $k\in\Omega^{(l)}$ is to the query $q$. While $\rel(k)$ might be given as input in information retrieval datasets \cite{DBLP:journals/corr/QinL13,chapelle2011yahoo}, we need to define it based on the hierarchical tree in our case. We want to enforce the constraint that the relevance decreases when going up the tree, \ie $\rel(k)>\rel(k')$ for $k\in\Omega^{(l)}$, $k'\in\Omega^{(l')}$ and $l>l'$. To do so, we assign a total weight of $(l/L)^\alpha$ to each semantic level $l$, where $\alpha\in\mathbb{R}^+$ controls the decrease rate of similarity in the tree. For example for $L=3$ and $\alpha=1$, the total weights for each level are $1$, $\frac{2}{3}$, $\frac{1}{3}$ and $0$. The instance relevance $\rel(k)$ is normalized by the cardinal of $\Omega^{(l)}$:
\begin{equation}\label{eq:hierarchy_relevance}
    \rel(k) = \frac{(l/L)^\alpha}{|\Omega^{(l)}|} \; \text{if } k \in \Omega^{(l)}
\end{equation}

\medbreak
We set $\alpha=1$ in~\cref{eq:hierarchy_relevance} for the $\hap$ metric and in our main experiments. Setting $\alpha$ to larger values supports better performances on fine-grained levels as their relevances will \textcolor{black}{relatively} increase. This variant is discussed in \cref{sec:hierarchical_results}. Other definitions of the relevance are possible, \eg an interesting option for the relevance enables to recover a weighted sum of AP, denoted as $\sum w\AP:=\sum_{l=1}^L w_l \cdot \AP^{(l)}$ (\textcolor{black}{supplementary~\cref{sec:hap_and_weighted_ap}}), \ie the weighted sum of AP is a particular case of $\hap$.

\subsubsection{Hierarchical Average Precision Training for Pertinent Image Retrieval}\label{sec:happier}

We define our surrogate loss to optimize $\hap$:

\begin{equation}
    \mathcal{L}_{\text{Sup-}\hap} = 1 - \frac{1}{M} \sum_{i=1}^M \frac{1}{\sum\limits_{k\in\Omega_i^+}\rel(k)} \sum_{k\in\Omega_i^+} \frac{\hrank(k)}{\rank^+(k) + \rank^-_s(k)}
\end{equation}

Note that in the hierarchical case $\rank^-(k)$ is the number of instances of relevances $<\rel(k)$ meaning that it may contain images that are similar to some extent to the query. Finally our ranking loss, \textbf{H}ierarchical \textbf{A}verage \textbf{P}recision training for \textbf{P}ertinent \textbf{I}mag\textbf{E} \textbf{R}etrieval (HAPPIER), is obtained by adding $\Ldecom$:
%\vspace{-1em}
\begin{equation}
    \mathcal{L}_{\text{HAPPIER}} = (1-\lambda) \cdot \mathcal{L}_{\text{Sup-}\hap} + \lambda \cdot \Ldecom
\end{equation}

\subsection{Application to the NDCG}\label{sec:ndcg}

The NDCG~\cite{jarvelin2002cumulated,croft2010search} is a common metric in the information retrieval community. The NDCG is defined using a relevance that is not required to be binary:
\begin{align}
    & \text{DCG}_i = \sum_{k\in\Omega_i^+} \frac{\rel(k)}{\log_2(1+\rank(k))} \nonumber \\  
    & \text{iDCG}_i = \max_{\rank} \text{DCG}_i \nonumber \\ 
    & \text{NDCG} = \frac{1}{M} \sum_{i=1}^M \frac{\text{DCG}_i}{\text{iDCG}_i}
\end{align}

We choose the following relevance function for the NDCG: $\rel(k) = 2^l - 1, \; \text{if } k \in \Omega^{(l)}$.
% \begin{equation}
%     \rel(k) = 2^l - 1, \; \text{if } k \in \Omega^{(l)}
% \end{equation}
Using the exponentiation is a standard procedure in information retrieval~\cite{croft2010search} as it allows to put more emphasis on instances of higher relevance. We then use similarly to other rank losses our SupRank surrogate. We use it to approximate the DCG, and thus the NDCG:

\begin{align}
    & \text{DCG}_{i,s} = \sum_{k\in\Omega_i^+} \frac{\rel(k)}{\log_2(1+\rank^+(k)+\rank_s(k))} \nonumber \\ 
    & \LsupNDCG = 1 - \frac{1}{M} \sum_{i=1}^M \frac{\text{DCG}_{i,s}}{\text{iDCG}_i}
\end{align}

Note that once again our surrogate loss, $\LsupNDCG$, is an upper bound on the true loss $1-\text{NDCG}$. Finally our training loss is:

\begin{equation}
    \mathcal{L}_{\text{ROD-NDCG}} = (1-\lambda)\cdot\LsupNDCG + \lambda\cdot\Ldecom
\end{equation}

\begin{figure*}
    \begin{subfigure}[b]{0.4\textwidth}
        \centering
        \includegraphics[width=\linewidth]{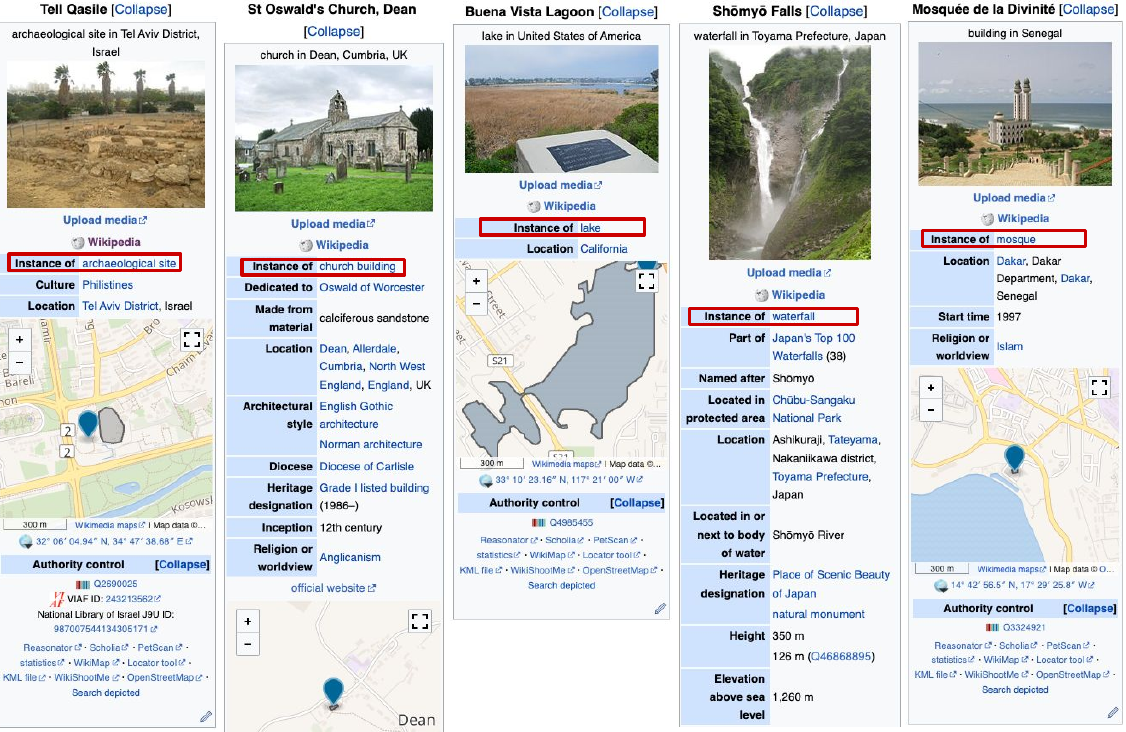}
        \caption{Screen captures of Wikimedia Commons webpages.}
        \label{fig:gldv2_scraping}
    \end{subfigure}
    \begin{subfigure}[b]{0.19\textwidth}
        \centering
        \includegraphics[width=0.7\linewidth]{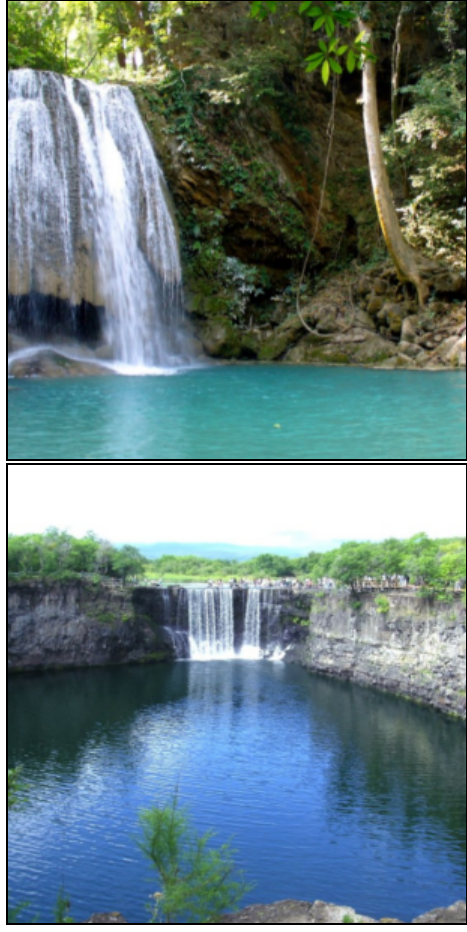}
        \caption{Waterfall.}
        \label{fig:gldv2_waterfall}
    \end{subfigure}
    \begin{subfigure}[b]{0.19\textwidth}
        \centering
        \includegraphics[width=0.7\linewidth]{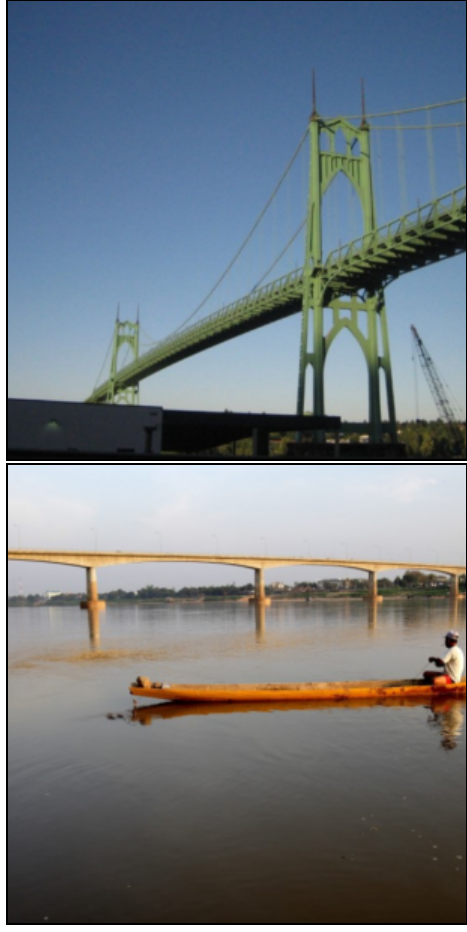}
        \caption{Bridge.}
        \label{fig:gldv2_bridge}
    \end{subfigure}
    \begin{subfigure}[b]{0.19\textwidth}
        \centering
        \includegraphics[width=0.7\linewidth]{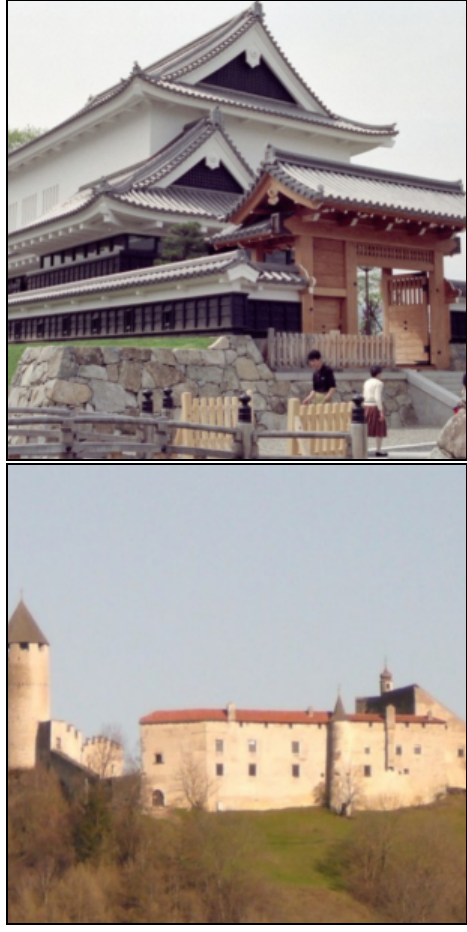}
        \caption{Castle.}
        \label{fig:gldv2_castle}
    \end{subfigure}
    \caption{\cref{fig:gldv2_scraping} depicts the ``Instance of'' (within red rectangles), from which we collect hierarchical landmark labels: \eg \emph{lake}, \emph{waterfall}, \emph{mosque}. \cref{fig:gldv2_waterfall,fig:gldv2_bridge,fig:gldv2_castle} illustrate some of the supercategories of our $\mathcal{H}$-GLDv2 dataset.}
    \vspace{-1.5em}
\end{figure*}
% \clearpage

\section{Hierarchical Landmark dataset}\label{sec:hierarchical_image_retrieval}

One of the most popular domains for image retrieval research is that of human-made and natural landmarks \cite{Radenovic-CVPR18,weyand2020google,chen2011city,avrithis2010feature,torii201524}.
In this work, we introduce for the first time a hierarchical dataset in this domain: $\mathcal{H}\text{-GLDv2}$, building on top of the Google Landmarks Dataset v2 (GLDv2) \cite{weyand2020google}, which is the largest and most diverse landmark dataset.
In the following, we present our process to semi-automatically annotate GLDv2 with an initial scraping of hierarchical labels from Wikimedia Commons, and a 2-step post-processing of the supercategories. We illustrate some of the created groups in~\cref{fig:gldv2_waterfall,fig:gldv2_bridge,fig:gldv2_castle}.
These hierarchical labels are released under the CC BY 4.0 license.

\vspace{-1em}
\subsection{Scraping Wikimedia Commons}\label{sec:scrapping_gldv2}

The landmarks from GLDv2 are sourced from Wikimedia Commons, the world’s largest crowdsourced collection of landmark photos.
After careful inspection, we find that many of the landmarks in GLDv2 can be associated to supercategories by leveraging the ``Instance of'' annotations available in Wikimedia Commons -- see~\cref{fig:gldv2_scraping}. Out of the original $203k$ landmarks in GLDv2-train, we were able to scrape supercategories for $129.1k$.
For the $101k$ landmarks in GLDv2-index, we were able to scrape supercategories for $68.1k$.
A lightweight manual cleaning process was then applied to remove landmarks assigned to more than one supercategory and those with irrelevant supercategories (\eg, supercategories named ``Wikimedia category'' or ``Wikimedia disambiguation page'').
Approximately $0.25$\% of landmarks end up being removed in this process, leading to a total number of selected landmarks of $128.8k$ and $67.9k$ for the train and index dataset splits, respectively.
The number of unique scraped supercategories is $5.7k$.

\vspace{-1em}
\subsection{Post-processing supercategories}

The scraped supercategories are noisy and do not have the same level of granularity, \eg ``church building'' \vs ``church building (1172–1954)''. To mitigate this issue after the scraping we perform a two step post-processing to obtain the final supercategories.

\begin{enumerate}
    \item \textbf{K-means clustering:} We first encode all the labels using the \textsc{CLIP}~\cite{Radford2021LearningTV} textual encoder. We perform a k-means on the latent representations. This initial clustering allows to show different prominent categories, \eg ``Church'', ``Castle'' \etc 
    \item \textbf{Manual verification:} We manually assess the obtained clusters based on the scraped label names. We create semantic groups by dividing the k-means clusters into sub-clusters. This leads to $78$ supercategories that we further group into human-made and natural landmarks. Two expert annotators comprehensively reviewed the final clusters manually and filtered them to produce a high-quality dataset.
\end{enumerate}

\vspace{-1.2em}
\subsection{Discussion and limitations}

$\mathcal{H}$-GLDv2 is a large scale dataset we were thus not able to manually check all images. This leads to a dataset that can have some noise. We release along with $\mathcal{H}$-GLDv2 the scraped labels to allow further work on the ``supercategories''.
Another difficulty of $\mathcal{H}$-GLDv2 is the ambiguity of some supercategories. For instance, the bottom image of~\cref{fig:gldv2_bridge} is labeled as ``Bridge'', however it could be labeled as ``River'', another supercategory.
Finally, there is an imbalance between supercategories that comes from the classes represented in GLDv2~\cite{weyand2020google}. We report first results in~\cref{sec:h_gldv2_exp} of models trained on our $\mathcal{H}$-GLDv2 dataset.
% \clearpage
%\vspace{-1.2em}

\section{Experiments}\label{sec:experiments}

\subsection{Standard image retrieval.}\label{sec:fine_grained_experiments}

In this section we compare our methods on the standard image retrieval setup, \ie $\rel(x_i,x_j)\in\{0,1\}$, and report fine-grained metrics. We use publicly available implementations of all baselines and run all experiments under the same settings. We use a ResNet-50 backbone with average pooling, a normalization layer without affine parameters and a projection head that reduces the dimension from $2048$ to $512$. We use a batch size of $256$ by sampling 4 images per class and the hierarchical samplig of~\cite{fastap} for SOP, with resolution $224\times224$, standard data augmentation (random resize crop, horizontal flipping), the Adam optimizer (with learning rate of $5\cdot10^{-5}$ on SOP and $1\cdot10^{-5}$ on iNaturalist, with cosine decay) and train for 100 epochs.

\subsubsection{Comparison to AP approximations}
In \cref{tab:compa_ranking_losses}, we compare ROADMAP to AP loss approximations including soft-binning approaches Fast-AP~\cite{fastap} and SoftBin-AP~\cite{revaud2019learning}, the generic solver BlackBox-AP~\cite{blackboxap}, and the smooth rank approximation~\cite{smoothap}. We observe that ROADMAP outperforms all the current AP approximations by a large margin. The gains are especially pronounced on the large-scale dataset iNaturalist.

\begin{table}[h]
    \caption{Comparison between ROADMAP and state-of-the-art AP ranking based methods.}
    \vspace{-0.5\intextsep}
        \label{tab:compa_ranking_losses}
        \centering
    \begin{tabularx}{0.5\textwidth}{ l YYYY}
        \toprule
         &  \multicolumn{2}{c}{SOP} & \multicolumn{2}{c}{iNaturalist} \\
         \midrule
         Method & R@1 & mAP@R & R@1 & mAP@R \\
         \midrule
         Fast-AP \footnotesize\cite{fastap} & 77.8 & 50.5 & 59.9 & 24.0  \\
         SoftBin-AP \footnotesize\cite{revaud2019learning} & 79.7 & 52.7 & 63.6 & 25.4 \\
         BlackBox-AP \footnotesize\cite{blackboxap} & 80.0 & 53.1 & 52.3 & 15.2  \\
         Smooth-AP \footnotesize\cite{smoothap} & 80.9 & 54.3 & 67.3 & 26.5 \\
         \midrule
         \textbf{ROADMAP} & \textbf{81.9} & \textbf{55.7} & \textbf{71.8} & \textbf{29.5} \\
         \bottomrule
    \end{tabularx}
    \vspace{-1em}
\end{table}

\subsubsection{Ablation study.}\label{sec:ablation_study}
To investigate more in-depth the impact of the two components of our framework, we perform ablation studies in \cref{tab:ablation_study}. We show the improvements against Smooth-AP~\cite{smoothap} and Smooth-R@k~\cite{patel2022recall} when replacing the sigmoid by SupRank~\cref{eq:average_precision_with_ranks}, and the use of $\Labs$~\cref{eq:labs} or $\Ldecom$~\cref{eq:cluster_loss}. We can see that both $\LsupAP$ and $\mathcal{L}_\text{Sup-R@k}$ consistently improve performances over the baselines, +0.5pt mAP@R on SOP and +1pt mAP@R on iNaturalist for both Sup-AP and Sup-R@k. Both $\Labs$ and $\Ldecom$ improve over the smooth surrogates, with strong gains on iNaturalist, \eg $\Ldecom$ improves by +2.9pt R@1 over Sup-AP and +3.7pt R@1 over Sup-R@k. This is because the batch vs. dataset size ratio $\frac{B}{N}$ is tiny ($\sim8\cdot10^{-4}\ll 1$), making the decomposability gap in \cref{eq:decomposability-gap} huge. On SOP $\Labs$ and $\Ldecom$ work similarly, however on iNat $\Ldecom$ performs far better than $\Labs$. In the following we choose to keep only $\Ldecom$.

\begin{table}[h]
    \vspace{-1em}
        \caption{Ablation study of the two components of our framework.}
        \vspace{-0.5\intextsep}
     \label{tab:ablation_study}
    \centering
    \begin{tabularx}{0.5\textwidth}{l YY YY YY}
        \toprule
         & & &  \multicolumn{2}{c}{SOP} & \multicolumn{2}{c}{iNaturalist}\\
         \midrule
         Method & rank  & $DG$ & {\scriptsize R@1} & {\scriptsize mAP@R} & {\scriptsize R@1} & {\scriptsize mAP@R} \\
         \midrule
         Smooth-AP &  sigmoid &  \xmark & 80.9 & 54.3 & 67.3 & 26.5 \\
         Sup-AP & SupRank &  \xmark & 81.2 & 54.8 & 68.9 & 27.5 \\
         \multirow{2}{*}{ROADMAP} &  \multirow{2}{*}{SupRank} &  $\Labs$ & 81.7 & \textbf{55.7} & 69.1 & 27.6 \\
          &  &  $\Ldecom$ & \textbf{81.9} & \textbf{55.7} & \textbf{71.8} & \textbf{29.5} \\
         \midrule
         \midrule
         Smooth-R@k &  sigmoid &  \xmark & 80.5 & 53.7 & 66.4 & 25.5 \\
         Sup-R@k & SupRank &  \xmark & 80.7 & 54.2 & 68.2 & 26.4 \\
         \multirow{2}{*}{ROD-R@k} & \multirow{2}{*}{SupRank} &  $\Labs$ & \textbf{82.4} & \textbf{56.6} & 69.3 & 27.0 \\
          &  &  $\Ldecom$ & {81.9} & {55.8} & \textbf{71.9} & \textbf{29.8} \\
         \bottomrule
    \end{tabularx}
    \vspace{-1.2em}
\end{table}

\subsubsection{Analysis on decomposability}

The decomposability gap depends on the batch size~\cref{eq:decomposability-gap}. To illustrate this we monitor on~\cref{fig:relative_increase} the relative improvement when adding $\Ldecom$ to $\LsupAP$ as the batch size decreases. We can see that the relative improvement becomes larger as the batch size gets smaller.
This confirms our intuition that the decomposability loss $\Ldecom$ has a stronger effect on smaller batch sizes, for which the AP estimation is noisier and $DG$ larger. This is critical on the large-scale dataset iNaturalist where the batch AP on usual batch sizes is a very poor approximation of the global AP.

\definecolor{redsop}{RGB}{250,164,146}
\definecolor{greeninat}{RGB}{198,250,146}
\definecolor{bluecub}{RGB}{146,225,250}

\pgfplotstableread[row sep=\\,col sep=&]{
    interval & cub & sop & inat \\
    32     & 3.4 & 5.5 & 22.8 \\
    64     & 2.6 & 2.5 & 13.1  \\
    128    & 2.4 & 1.4 & 6.7 \\
    256    & 2.3 & 0.7 & 4.1 \\
    384    & 2.0 & 0.7 & 3.5 \\
    }\mydata

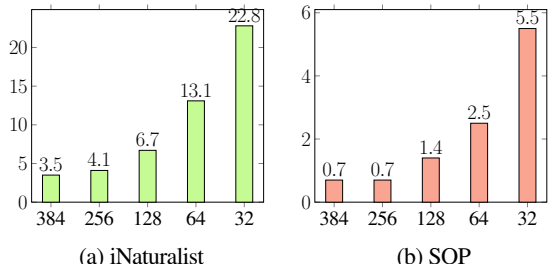
\begin{figure}[h!]
    \vspace{-1em}
    \centering
    \begin{subfigure}[t]{0.2\textwidth}
    \begin{tikzpicture}[scale=0.45]
        \begin{axis}[
                ybar,
                font=\LARGE,
                bar width=.5cm,
                legend style={at={(0.5,1)},
                    anchor=north,legend columns=-1},
                symbolic x coords={32,64,128,256,384},
                xtick=data,
                nodes near coords,
                nodes near coords align={vertical},
                ymin=0,ymax=24.9,
                ylabel={},
                x dir=reverse
            ]
            \addplot[fill=greeninat] table[x=interval,y=inat]{\mydata};
            % \legend{}
        \end{axis}
    \end{tikzpicture}
    \caption{iNaturalist}
    \label{fig:relative_increase_inat}
    \end{subfigure}
    ~
    \begin{subfigure}[t]{0.2\textwidth}
    \begin{tikzpicture}[scale=0.45]
        \begin{axis}[
                ybar,
                font=\LARGE,
                bar width=.5cm,
                legend style={at={(0.5,1)},
                    anchor=north,legend columns=-1},
                symbolic x coords={32,64,128,256,384},
                xtick=data,
                nodes near coords,
                nodes near coords align={vertical},
                ymin=0,ymax=6.1,
                ylabel={},
                x dir=reverse
            ]
            \addplot[fill=redsop] table[x=interval,y=sop]{\mydata};
            % \legend{}
        \end{axis}
    \end{tikzpicture}
    \caption{SOP}
    \label{fig:relative_increase_sop}
    \end{subfigure}

    \vspace{-0.4\intextsep}
    \caption{Relative increase of mAP@R \vs batch size when adding $\Ldecom$ to $\LsupAP$.}
    \label{fig:relative_increase}
    \vspace{-1.5em}
\end{figure}

In~\cref{tab:compa_xbm} we compare ROADMAP to the cross-batch memory~\cite{xbm} (XBM) which is used reduce the gap between batch-AP and global AP. We use XBM with a batch size of 128 and store all the dataset, and use the setup described previously otherwise. ROADMAP outperforms XBM both on SOP and iNaturalist with gains more pronounced on iNaturalist with +12.5pt R@1 and +11 mAP@R. $\Ldecom$ allows us to train models even with smaller batches.

\begin{table}[ht]
    % \vspace{-0.8\intextsep}
    \caption{Comparison between XBM~\cite{xbm} and ROADMAP equiped with memory.}
    \vspace{-0.5\intextsep}
        \label{tab:compa_xbm}
        \centering
    \begin{tabularx}{0.5\textwidth}{ l YYYY}
        \toprule
         &  \multicolumn{2}{c}{SOP} & \multicolumn{2}{c}{iNaturalist} \\
         \midrule
         Method & R@1 & mAP@R & R@1 & mAP@R \\
         \midrule
         XBM \footnotesize\cite{xbm} & 80.6 & 54.9 & 59.3 & 18.5 \\
         \textbf{ROADMAP} & \textbf{81.9} & \textbf{55.7} & \textbf{71.8} & \textbf{29.5} \\
         \bottomrule
    \end{tabularx}
    \vspace{-1em}
\end{table}

\begin{table*}[t]
    \caption{Comparison of state-of-the-art performances on R@K from the literature on SOP, CUB, and iNaturalist with the proposed ROADMAP. Except for the ViT categories, all methods rely on a standard convolutional backbone (generally ResNet-50).}
    \vspace{-0.5\intextsep}
    \setlength\tabcolsep{3pt}
    \label{tab:general_results_sop}
    \begin{tabularx}{\textwidth}{ l l>{\small}c |YYY|YYYY|YYYY }
        \toprule
        & & & \multicolumn{3}{c}{SOP} & \multicolumn{4}{c}{CUB} & \multicolumn{4}{c}{iNaturalist}\\
        %\hline
        & Method & dim & 1 & 10 & 100 & 1 & 2 & 4 & 8 & 1 & 4 & 16 & 32\\
        \midrule
        \multirow{6}{*}{\rotatebox[origin=c]{90}{Metric learning}}
        & Triplet SH \cite{wu2017sampling} & 512 & 72.7 & 86.2 & 93.8 & 63.6 & 74.4 & 83.1 & 90.0 & 58.1 & 75.5 & 86.8 & 90.7\\
        & MS \cite{multi_similarity} & 512 & 78.2 & 90.5 & 96.0 & 65.7 & 77.0 & 86.3 & 91.2 &-&-&-&-\\
        & SEC \cite{spherical_embedding} & 512 & 78.7 & 90.8 & 96.6 & 68.8 & 79.4 & 87.2 & 92.5 &-&-&-&-\\
        & HORDE \cite{horde} & 512 & 80.1 & 91.3 & 96.2 & 66.8 & 77.4 & 85.1 & 91.0 &-&-&-&-\\
        & XBM \cite{xbm} & 128 & 80.6 & 91.6 & 96.2 & 65.8 & 75.9 & 84.0 & 89.9 &-&-&-&-\\
        & Triplet SCT \cite{xuan2020hard} & 512/64 & 81.9 & 92.6 & 96.8 & 57.7 & 69.8 & 79.6 & 87.0 &-&-&-&-\\
        \midrule
        \multirow{7}{*}{\rotatebox[origin=c]{90}{Classification}}
        & ProxyNCA \cite{proxynca} & 512 & 73.7 & - & - & 49.2 & 61.9 & 67.9 & 72.4 & 61.6 & 77.4 & 87.0 & 90.6 \\
        & ProxyGML \cite{fewer_is_more} & 512 & 78.0 & 90.6 & 96.2 & 66.6 & 77.6 & 86.4 & - &-&-&-&-\\
        & NSoftmax \cite{norm_softmax} & 512 & 78.2 & 90.6 & 96.2 & 61.3 & 73.9 & 83.5 & 90.0 &-&-&-&-\\
        & NSoftmax \cite{norm_softmax} & 2048 & 79.5 & 91.5 & 96.7 & 65.3 & 76.7 & 85.4 & 91.8 &-&-&-&-\\
        & Cross-Entropy \cite{unifying_mi} & 2048 & 81.1 & 91.7 & 96.3 & 69.2 & 79.2 & 86.9 & 91.6 &-&-&-&-\\
        & ProxyNCA++ \cite{proxynca++} & 512 & 80.7 & 92.0 & 96.7 & 69.0 & {79.8} & {87.3} & {92.7} &-&-&-&-\\
        & ProxyNCA++ \cite{proxynca++} & 2048 & 81.4 & 92.4 & 96.9 & \textbf{72.2} & \textbf{82.0} & \textbf{89.2} & \textbf{93.5} &-&-&-&-\\
        \midrule
        \multirow{6}{*}{\rotatebox[origin=c]{90}{Ranking}}
        & FastAP \cite{fastap} & 512 & 76.4 & 89.0 & 95.1 & -&-&-&- & 60.6 & 77.0 & 87.2 & 90.6\\
        & BlackBox \cite{blackboxap} & 512 & 78.6 & 90.5 & 96.0 & 64.0 & 75.3 & 84.1 & 90.6 & 62.9 & 79.4 & 88.7 & 91.7\\
        & SmoothAP \cite{smoothap} & 512 & 80.1 & 91.5 & 96.6 &-&-&-&- & 67.2 & 81.8 & 90.3 & 93.1\\ 
        & R@k \cite{patel2022recall} & 512 & 82.8 & 92.9 & 97.0 &-&-&-&-& 71.2 & 84.0 & 91.3 & 93.6 \\
        & R@k + SiMix \cite{patel2022recall} & 512 & 82.1 & 92.8 & 97.0 &-&-&-&-& 71.8 & 84.7 & 91.9 & 94.3 \\
        & \textbf{ROADMAP (ours)} & 512 & \textbf{83.3} & \textbf{93.6} & \textbf{97.4} & {69.4} & 79
        4 & 87.2 & 92.1 & \textbf{73.1} & \textbf{85.7} & \textbf{92.7} & \textbf{94.8} \\
        \midrule
        \multirow{2}{*}{\rotatebox[origin=c]{90}{{\scriptsize DeiT-S}}}
        & IRT\textsubscript{R} \cite{transformer_ir} & 384 & 84.2 & 93.7 & 97.3 & 76.6 & 85.0 & 91.1 & 94.3 &-&-&-&-\\
        & \textbf{ROADMAP (ours)} & 384 & \textbf{85.2} & \textbf{94.5} & \textbf{97.9} & \textbf{77.6} & \textbf{86.2} & \textbf{91.6} & \textbf{95.0} & \textbf{74.7} & \textbf{86.9} & \textbf{93.4} & \textbf{95.4} \\
        \midrule
        \multirow{2}{*}{\rotatebox[origin=c]{90}{{\scriptsize ViT-B}}}
        & R@k + SiMix~\cite{patel2022recall} & 512 & 88.0 & 96.1 & 98.6 &-&-&-&-& 83.9 & 92.1 & 95.9 & 97.2 \\
        & \textbf{ROADMAP (ours)} & 512 & \textbf{88.4} & \textbf{96.4} & \textbf{98.7} & \textbf{86.8} & \textbf{91.7} & \textbf{94.6} & \textbf{96.5} & \textbf{85.1} & \textbf{93.0} & \textbf{96.6} & \textbf{97.7} \\
        \bottomrule
    \end{tabularx}
    \vspace{-1em}
\end{table*}

\subsubsection{ROADMAP hyper-parameters}\label{sec:roadmap_hyp}

\begin{figure}[h!]
\vspace{-1em}
    \centering
    \begin{subfigure}[t]{0.2\textwidth}
        \begin{tikzpicture}[scale=0.45]
        \begin{axis}[
            title={},
            xlabel={},
            ylabel={mAP@R},
            xmin=-0.03, xmax=1,
            ymin=68, ymax=72.2,
            xtick={0,0.1,0.3,0.5,0.7,0.9,1.0},
            ytick={},
            legend pos=south east,
            ymajorgrids=true,
            grid style=dashed,
            % x dir=reverse
            font=\LARGE,
        ]
        \addplot[
            line width=2,
            color=blue,
            mark=x,
            mark size=5,
            ]
            coordinates {
            (0.0,68.93)(0.1,71.76)(0.3,72.07)(0.5,71.96)(0.7,71.72)(0.9,71.45)(1.0,70.2)
            };
        \end{axis}
        \end{tikzpicture}
      \caption{R@1 \vs $\lambda$ for $\LROADMAP$}
      \label{fig:lambda_roadmap}
    \end{subfigure}
    ~
    \begin{subfigure}[t]{0.2\textwidth}
        \begin{tikzpicture}[scale=0.45]
        \begin{axis}[
            title={},
            xlabel={},
            ylabel={},
            xmin=0.07, xmax=10001,
            ymin=26, ymax=27.6,
            xtick={0.1,1.0,10.0,100,1000,10000},
            ytick={},
            legend pos=south east,
            ymajorgrids=true,
            grid style=dashed,
            xmode=log,
            font=\LARGE,
            % x dir=reverse
        ]
        \addplot[
            line width=2,
            color=blue,
            mark=x,
            mark size=5,
            ]
            coordinates {
            (0.1,26.52)(1,26.67)(10.0,27.19)(50.0,27.42)(100,27.48)(500,27.43)(1000,27.34)(10000,26.51)
            };
        \end{axis}
        \end{tikzpicture}
        \caption{mAP@R \vs $\rho$ for $\LsupAP$}
        \label{fig:rho_supap}
    \end{subfigure}

    \vspace{-0.4\intextsep}
    \caption{Robustness to hyper-parameters on iNaturalist.}
    \label{fig:supap_hyperparameters}
    \vspace{-1.5em}
\end{figure}
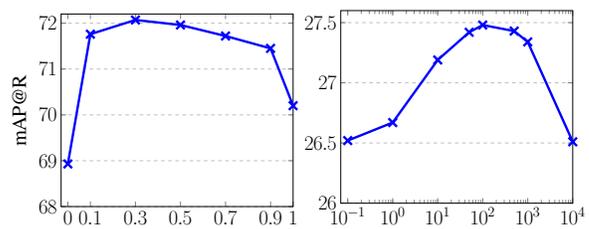

We demonstrate the robustness of our framework to hyper-parameters in~\cref{fig:supap_hyperparameters}.
Firstly, \cref{fig:lambda_roadmap} illustrates the complementarity between the two terms of $\LROADMAP$. For $0<\lambda<1$, $\LROADMAP$ outperforms both $\LsupAP$ and $\Ldecom$. While we use $\lambda=0.1$ in our experiments, hyper-parameter tuning could yield better results, \eg with $\lambda=0.3$ $\LROADMAP$ has 72.1 R@1 \vs 71.8 R@1 reported in~\cref{tab:compa_ranking_losses}. Secondly \cref{fig:rho_supap} shows the influence of the slope $\rho$ that controls the linear regime in $H^-$. As shown in \cref{fig:rho_supap}, the improvement is important and stable in $[10, 100]$. Note that $\rho>1$ already improves the results compared to $\rho=0$ in \cite{smoothap}. There is a decrease when $\rho \gg 10^3$ probably due to the high gradient that takes over the signal for correctly ranked samples.

\vspace{-2em}
\subsection{Comparison to state-of-the-art}\label{sec:sota}

In this section, we compare our AP approximation method, ROADMAP, to state-of-the-art methods, on SOP, CUB, and iNaturalist. We use ROADMAP with a memory~\cite{xbm} to virtually increase the batch size. Note that using batch memory is less computationally expensive than methods such as~\cite{patel2022recall} which trade computational time for memory footprint by using two forward passes. We apply ROADMAP on both a convolutional backbone, ResNet-50 with GeM pooling~\cite{Radenovic-CVPR18} and layer normalization, and Vision transformer models~\cite{vit}, DeiT-S~\cite{deit} (Imagenet-1k pre-trained as in~\cite{transformer_ir}) and ViT-B (Imagenet21k pre-trained as in~\cite{patel2022recall}). For convolutional backbones, we choose to keep the standard images of size $224\times224$ for both training and inference on SOP and iNaturalist, and use more recent settings~\cite{proxynca++,xuan2020hard} for CUB and use images of size $256\times256$. Vision transformers experiments use images of size $224\times224$.

In~\cref{tab:general_results_sop}, using convolutional backbones, ROADMAP outperforms most state-of-the-art methods when evaluated at different (standard) R@k. As ROADMAP optimizes directly the evaluation metrics, it outperforms metric learning and classification-based methods, \eg +1.4pt R@1 on SOP compared to Triplet SCT~\cite{xuan2020hard} or +1.9pt R@1 on SOP \vs ProxyNCA++~\cite{proxynca++}. ROADMAP also outperforms R@k~\cite{patel2022recall} with +1.2pt R@1 on SOP and +1.3pt R@1 on iNaturalist. This is impressive as R@k~\cite{patel2022recall} uses a strong setup \ie a batch size of $4096$ and Similarity mixup.
On the small-scale dataset CUB, our method is competitive with methods such as ProxyNCA++ with the same embedding size of 512

Finally, we show that ROADMAP also improves Vision Transformers for image retrieval. With DeiT-S, ROADMAP outperforms~\cite{transformer_ir} on both SOP and CUB by +1pt R@1, this again shows the interest of directly optimizing the metrics rather than the pair loss of~\cite{xbm} used in~\cite{transformer_ir}. With ViT-B, ROADMAP outperforms~\cite{patel2022recall} by +0.4pt R@1 and +1.2pt R@1 on SOP and iNaturalist respectively. We attribute this to the fact that our loss is an actual upper bound of the metric, in addition to our decomposability loss.

\begin{table*}[ht]
    \setlength\tabcolsep{0.5pt}
    \caption{Comparison of HAPPIER on SOP and iNat-base/full. Best results in \textbf{bold}, second best \underline{underlined}.
    }
    \vspace{-0.5\intextsep}
    \label{tab:main_sop_inat} 
    \centering
    \begin{tabularx}{\textwidth}{ l l YYYYY | YYYYY | YYYYY}
        \toprule
        & \multirow{2}{*}{Method}  & \multicolumn{5}{c|}{SOP}  & \multicolumn{5}{c|}{iNat-base} & \multicolumn{5}{c}{iNat-full} \\
        \cmidrule{3-17}
        && \footnotesize{R@1} & \footnotesize{AP} & \footnotesize{$\mathcal{H}$-AP} & \footnotesize{ASI} & \footnotesize{NDCG} & \footnotesize{R@1}  &\footnotesize{AP} & \footnotesize{$\mathcal{H}$-AP} & \footnotesize{ASI} & \footnotesize{NDCG} & \footnotesize{R@1} & \footnotesize{AP} & \footnotesize{$\mathcal{H}$-AP} & \footnotesize{ASI} & \footnotesize{NDCG} \\
         \midrule
         \multirow{4}{*}{\rotatebox[origin=c]{90}{Fine}}
         & Triplet SH~\cite{wu2017sampling} & 79.8 & 59.6 & 42.2 & 22.4 & 78.8 & 66.3 & 33.3 & 39.5 & 63.7 & 91.5 & 66.3 & 33.3 & 36.1 & 59.2 & 89.8 \\
         & NSM~\cite{norm_softmax} & 81.3 & 61.3 & 42.8 & 21.1 & 78.3 & 70.2 & \underline{37.6} & 38.0 & 51.6 & 88.9 & 70.2 & \textbf{37.6} & 33.3 & 51.7 & 88.2 \\
         & NCA++~\cite{proxynca++} & \underline{81.4} & \underline{61.7} & 43.0 & 21.5 & 78.4 & 67.3 & 35.2 & 39.5 & 57.0 & 90.1 & 67.3 & 35.2 & 35.3 & 55.7 & 89.0  \\
         & \textcolor{black}{Smooth-AP~\cite{smoothap}} & 80.9 & 60.8 & 42.9 & 20.6 & 78.2 & 67.3 & 35.2 & 41.3 & 64.2 & 91.9 & 67.3 & 35.2 & 37.2 & 60.1 & 90.1 \\
         \midrule
         \multirow{7}{*}{\rotatebox[origin=c]{90}{Hier.}}
         & $\Sigma\text{TL}_{\text{SH}}$~\cite{wu2017sampling} & 78.3 & 57.6 & {53.1} & 53.3 & {89.2} & 54.7 & 21.3 & 44.0 & 87.4 & 96.4 & 52.9 & 19.7 & 39.9 & \underline{85.5} & 92.0 \\ 
         & $\Sigma$NSM~\cite{norm_softmax} & 79.4 & 58.4 & 50.4 & 49.7 & 87.0 & 69.5 & 37.5 & 47.9 & 75.8 & 94.4 & 67.2 & 36.1 & \underline{46.9} & 74.2 & \textbf{93.8}  \\
         & $\Sigma$NCA++~\cite{proxynca++} & 76.3 & 54.5 & 49.5 & 52.8 & 87.8 & 64.2 & 35.4 & 48.9 & 78.7 & 95.0 & 67.4 & 36.3 & 44.7 & 74.3 & 92.6 \\
         & CSL~\cite{sun2021dynamic} & 79.4 & 58.0 & 52.8 & \underline{57.9} & 88.1 & 62.9 & 30.2 & {50.1} & \textbf{89.3} & \underline{96.7} & 59.9 & 30.4 & 45.1 & 84.9 & 93.0 \\ 
        %  \midrule
        \cmidrule{2-17}
        & \textbf{ROD-NDCG (ours)} & 80.5 & 59.6 & \underline{58.3} & \underline{65.0} & \underline{91.1} & \underline{70.7} & 35.9 & \underline{53.1} & 87.8 & \underline{96.6} & \underline{71.2} & 36.7 & 44.8 & 81.1 & \underline{93.1} \\
        \cmidrule{2-17}
         & \textbf{HAPPIER (ours)} & 81.0 & 60.4 & \textbf{59.4} & \textbf{65.9} & \textbf{91.5} & \underline{70.7} & 36.7 & \textbf{54.3} & \textbf{89.3} & \textbf{96.9} & 70.2 & 36.0 & \textbf{47.9} & \textbf{87.2} & \textbf{93.8} \\
         & \textbf{HAPPIER\textsubscript{F} (ours)} & \textbf{81.8} & \textbf{62.2} & 52.0 & 45.9 & 86.5 & \textbf{71.6} & \textbf{37.8} & 43.2 & 87.0 & 96.6 & \textbf{71.4} & \textbf{37.6} & 40.1 & 80.0 & 93.5 \\
        \bottomrule
    \end{tabularx}
\end{table*}

\begin{table*}[ht]
    \caption{Comparison of HAPPIER \vs fine-grained methods and CSL on iNat-full. Metrics are reported for all 7 semantic levels.}
    \vspace{-0.5\intextsep}
    \label{tab:main_detail_inat_full} 
    \centering
    \begin{tabularx}{\textwidth}{l l YY Y Y Y Y Y Y}
        \toprule
        & \multirow{2}{*}{Method} & \multicolumn{2}{c}{\scriptsize{Species}} & \multicolumn{1}{c}{\scriptsize{Genus}} & \multicolumn{1}{c}{\scriptsize{Family}} & \multicolumn{1}{c}{\scriptsize{Order}} & \multicolumn{1}{c}{\scriptsize{Class}} & \multicolumn{1}{c}{\scriptsize{Phylum}} & \multicolumn{1}{c}{\scriptsize{Kingdom}} 
        \\
         && \raone & \sap & \sap & \sap & \sap & \sap & \sap & \sap \\
         \midrule
         \multirow{4}{*}{\rotatebox[origin=c]{90}{Fine}}
         & $\text{TL}_{\text{SH}}$~\cite{wu2017sampling} & 66.3 & 33.3 & 34.2 & 32.3 & 35.4 & 48.5 & 54.6 & 68.4 \\
         & NSM~\cite{norm_softmax} & \underline{70.2} & \textbf{37.6} & \underline{38.0} & 31.4& 28.6 & 36.6 & 43.9 & 63.0 \\
         & NCA++~\cite{proxynca++} & 67.3 & 37.0 & 37.9 & 33.0 & 32.3 & 41.9 & 48.4 & 66.1 \\
         & \textcolor{black}{Smooth-AP~\cite{smoothap}} & 67.3 & 35.2 & 36.3 & 33.5 & 35.0 & 49.3 & 55.8 & 69.9 \\
         \midrule
         \multirow{3}{*}{\rotatebox[origin=c]{90}{Hier.}}
         & CSL~\cite{sun2021dynamic} & 59.9 & 30.4 & 32.4 & 36.2 & 50.7 & \underline{81.0} & \underline{87.4} & \underline{91.3} \\
        \cmidrule {2-10}
         & \textbf{HAPPIER (ours)} & \underline{70.2} & 36.0 & 37.0 & \underline{38.0} & \textbf{51.9} & \textbf{81.3} & \textbf{89.1} & \textbf{94.4} \\
         & \textbf{$\text{HAPPIER}_{\text{F}}$ (ours)} & \textbf{70.8} & \textbf{37.6} & \textbf{38.2} & \textbf{38.8} & \underline{50.9} & 76.1 & 82.2 & 83.1 \\
         \bottomrule
    \end{tabularx}
    \vspace{-1.5em}
\end{table*}

\vspace{-1em}
\subsection{Hierarchical Results}\label{sec:hierarchical_results}

In this section, we show results on the hierarchical settings and use the labels as described in the additional context of~\cref{sec:hierarchical_image_retrieval}. We report results using the experimental setting of~\cref{sec:fine_grained_experiments}. Additionally to hierarchical metrics NDCG and $\hap$ we report ASI which is defined in~\cref{sec:sup_asi}.

On \cref{tab:main_sop_inat}, we show that HAPPIER significantly outperforms methods trained on the fine-grained level only, with a gain on $\hap$ over the best performing methods of +16.4pt  $\hap$ on SOP, \textcolor{black}{+13pt}%+14.1pt
~on iNat-base and \textcolor{black}{10.7pt}% +13.8pt
~on iNat-full. HAPPIER also exhibits significant gains compared to hierarchical methods. On $\hap$, HAPPIER has important gains on all datasets (\eg +6.3pt on SOP, +4.2pt on iNat-base over the best competitor), but also on ASI %\andre{what is ASI?}
and NDCG. This shows the strong generalization of the method on standard metrics. Compared to the recent CSL loss~\cite{sun2021dynamic}, we observe a consistent gain over all metrics and datasets, \eg +6pt on $\hap$, +8pt on ASI and +2.6pts on NDCG on SOP. This shows the benefits of optimizing a well-behaved hierarchical metric compared to an ad-hoc proxy method.

Furthermore we can see that HAPPIER performs on-par to the best methods for standard image retrieval when considering fine-grained metrics. HAPPIER has 81.0 R@1 on SOP \vs 81.4 R@1 for NCA++, and even performs slightly better on iNat-base with 70.7 R@1 \vs 70.2 R@1 for NSM. Finally our variant HAPPIER\textsubscript{F} for $\alpha>1$~\cref{seq:relevance}, performs as expected ($\alpha$ is 5 on SOP and 3 on iNat-base/full): it is a strong method for fine-grained image retrieval, and still outperforms standard methods on hierarchical metrics.

\subsubsection{Detailed evaluation} HAPPIER performs well on the overall hierarchical metrics because it performs well on \emph{all} the hierarchical level. We illustrate this on~\cref{tab:main_detail_inat_full} which reports the different methods' performances on all semantic hierarchy levels on iNat-full. We evaluate HAPPIER and $\text{HAPPIER}_{\text{F}}$. HAPPIER optimizes the overall hierarchical performance, while $\text{HAPPIER}_{\text{F}}$ is meant to be optimal at the fine-grained level without sacrificing coarser levels. The satisfactory behavior and the two optimal regimes of HAPPIER and $\text{HAPPIER}_{\text{F}}$ are confirmed on iNat-full: HAPPIER gives the best results on coarser levels (from ``Class''), while being very close to the best results on finer ones. $\text{HAPPIER}_{\text{F}}$ gives the best results at the finest levels, even outperforming very competitive fine-grained baselines. HAPPIER also outperforms CSL~\cite{sun2021dynamic} on all semantic levels, \eg +5pt on the fine-grained AP (``Species'') and +3pt on the coarsest AP (``Kingdom''). We show the detailed evaluation on SOP and iNat-base in~\cref{sec:sup_detailed_eval}.

\subsubsection{Model analysis}

\begin{figure}[ht]
     \centering
        
     \begin{subfigure}[t]{0.23\textwidth}
     % \vspace{0pt}
         \includegraphics[width=\textwidth]{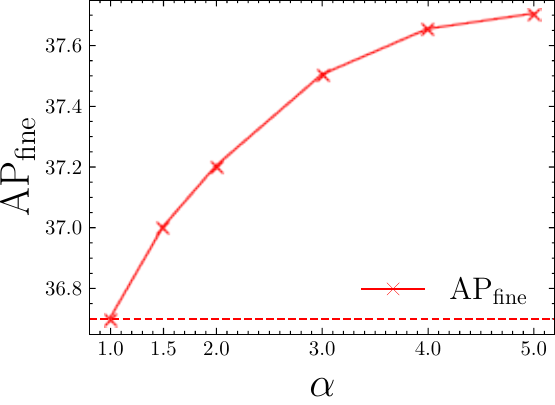}
         \caption{$\AP_{\text{fine}}$ vs $\alpha$ in~\cref{eq:hierarchy_relevance}.
         }
         \label{fig:analysis_alpha}
     \end{subfigure}
     ~
     \begin{subfigure}[t]{0.23\textwidth}
     % \vspace{0pt}
     \includegraphics[width=\textwidth]{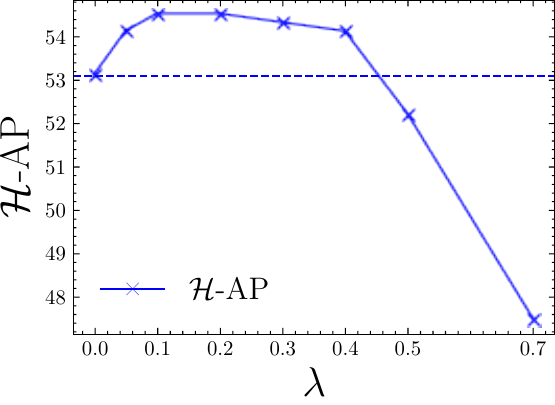}
     \caption{$\hap$ \vs $\lambda$ for $\mathcal{L}_{\text{HAPPIER}}$.
     }
     \label{fig:analysis_lambda}
     \end{subfigure}%

    \vspace{-0.4\intextsep}
    \caption{Impact on iNat-base of $\alpha$ in~\cref{eq:hierarchy_relevance} for setting the relevance of $\hap$ (a) and of the $\lambda$ hyper-parameter on HAPPIER results (b).}
    
\label{fig:analysis_alpha_lambda}
\vspace{-1em}
\end{figure}

We showcase the different behavior and the robustness of HAPPIER when changing the hyper-parameters. \cref{fig:analysis_alpha} studies the impact of $\alpha$ for setting the relevance in~\cref{eq:hierarchy_relevance}. $\alpha$ controls the balance between the relevance weight allocated to each levels. Increasing $\alpha$ puts more emphasis on the fine-grained levels, on the contrary diminishing its value will put an equal contribution to all levels. This is illustrated in~\cref{fig:analysis_alpha}: increasing $\alpha$ improves the AP at the fine-grained level on iNat-base. \cref{fig:analysis_alpha} shows that one can use $\alpha$ to obtain a range of performances for desired applications.

We measure the impact in \cref{fig:analysis_lambda} of $\lambda$ for weighting $\lhaps$ and $\lclust$ in HAPPIER: we observe a stable increase in $\hap$ with $0<\lambda<0.5$ compared to optimizing only $\lhaps$, while a drop in performance is observed for $\lambda>0.5$. This shows the complementarity of $\lhaps$ and $\Ldecom$, and how, when combined, HAPPIER reaches its best performance.

\subsubsection{Hierarchical landmark results}\label{sec:h_gldv2_exp}

In this section we report first results on our $\mathcal{H}$-GLDv2 dataset. We run all experiments under the same settings: we use a ResNet-101 with GeM pooling and initialize a linear projection with a PCA~\cite{revaud2019learning}. We use a batch size of 256 and train for $\sim55$k steps with Adam and a learning rate of $10^{-5}$ decayed using a cosine schedule. We report the mAP@100~\cite{weyand2020google}, and the hierarchical metrics $\hap$, ASI and NDCG.

\begin{table}[ht]
    \setlength\tabcolsep{0.5pt}
    \caption{Comparison of ROADMAP and HAPPIER \vs baselines on $\mathcal{H}$-GLDv2.
    }
    \vspace{-0.5\intextsep}
    \label{tab:landmark_retrieval}
    \centering
    % \begin{tabular}{ l cc cc cc}
    \begin{tabularx}{0.5\textwidth}{l YYYY}
        \toprule
         Method &  mAP@100 & $\hap$ & ASI & NDCG \\
         \midrule
         SoftBin~\cite{revaud2019learning} & 39.0 & 35.2 & 74.6 & 94.4 \\
         Smooth-AP~\cite{smoothap} & 42.5 & 37.3 & 76.9 & 94.7 \\
         R@k~\cite{patel2022recall} & 41.6 & 36.8 & 77.1 & 94.7 \\
         %\midrule
         \textbf{ROADMAP} & \underline{42.9} & 37.0 & 75.0 & 94.4 \\
         \midrule
         \midrule
         CSL~\cite{sun2021dynamic} & 37.5 & 36.2 & \textbf{85.4} & \textbf{95.7} \\
         \textbf{HAPPIER} & 41.6 & \textbf{38.8} & \underline{83.8} & \textbf{95.7}\\
         \textbf{HAPPIER\textsubscript{F}} & \textbf{43.7} & \underline{38.3} & 77.5 & 94.8 \\
         \bottomrule
    \end{tabularx}
    \vspace{-1.2em}
\end{table}

In~\cref{tab:landmark_retrieval} we report the results of ROADMAP and HAPPIER \vs other fine-grained methods and hierarchical methods. \cref{tab:landmark_retrieval} demonstrates once again the interest of our AP surrogate, ROADMAP and HAPPIER\textsubscript{F} perform the best on the fine-grained metric mAP@100. Furthermore HAPPIER has the best hierarchical results. It outperforms ROADMAP by +2.8pt $\hap$ and +8.8pt ASI. It also outperforms CSL by +2.6pt $\hap$.

\begin{figure*}[t]
% \vspace{-20pt}
\begin{minipage}[t]{0.3\textwidth}
    \begin{subfigure}[t]{\textwidth}
        \centering
        \includegraphics[width=0.7\textwidth]{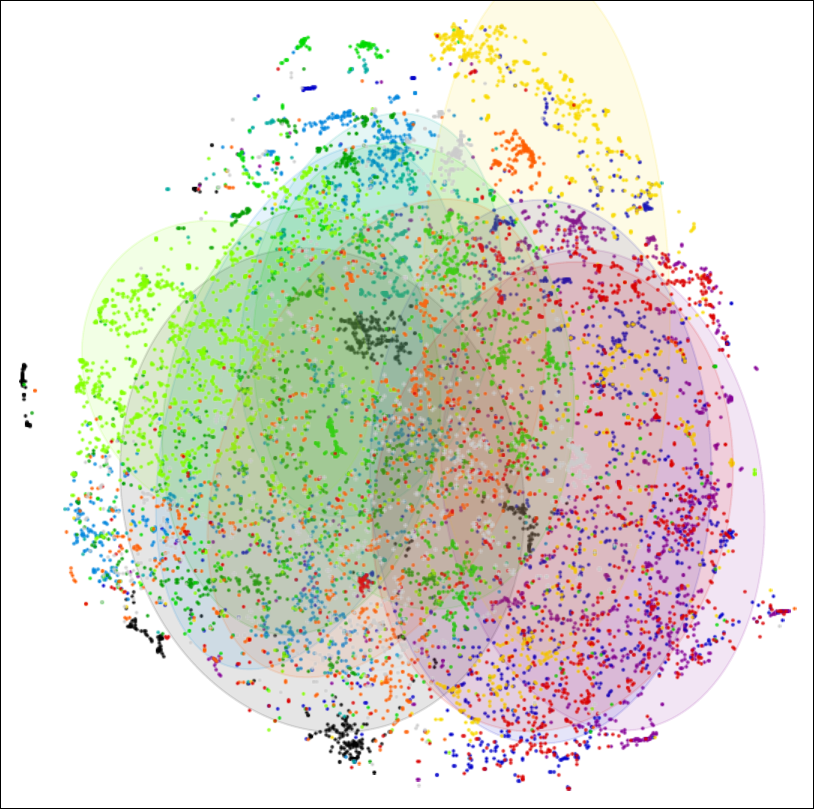}
        \caption{t-SNE visualization of a model trained on fine-grained labels with Smooth-AP~\cite{smoothap}.}
        \label{fig:tsne_baseline}
    \end{subfigure}
    ~
    \begin{subfigure}[t]{\textwidth}
    \centering
    \includegraphics[width=0.7\textwidth]{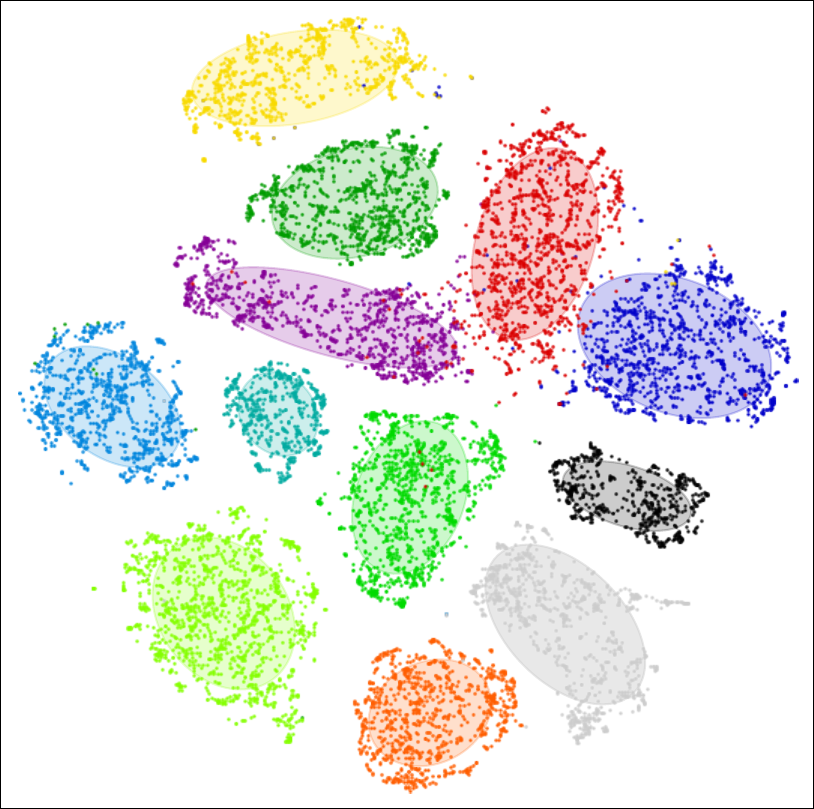}
    \caption{t-SNE visualization of a model trained with \textbf{HAPPIER}.}
    \label{fig:tsne_happier}
    \end{subfigure}
\end{minipage}%
\hfill
\begin{minipage}[t]{0.69\textwidth}
    \begin{subfigure}[t]{\textwidth}
        \centering
        \includegraphics[width=0.95\textwidth]{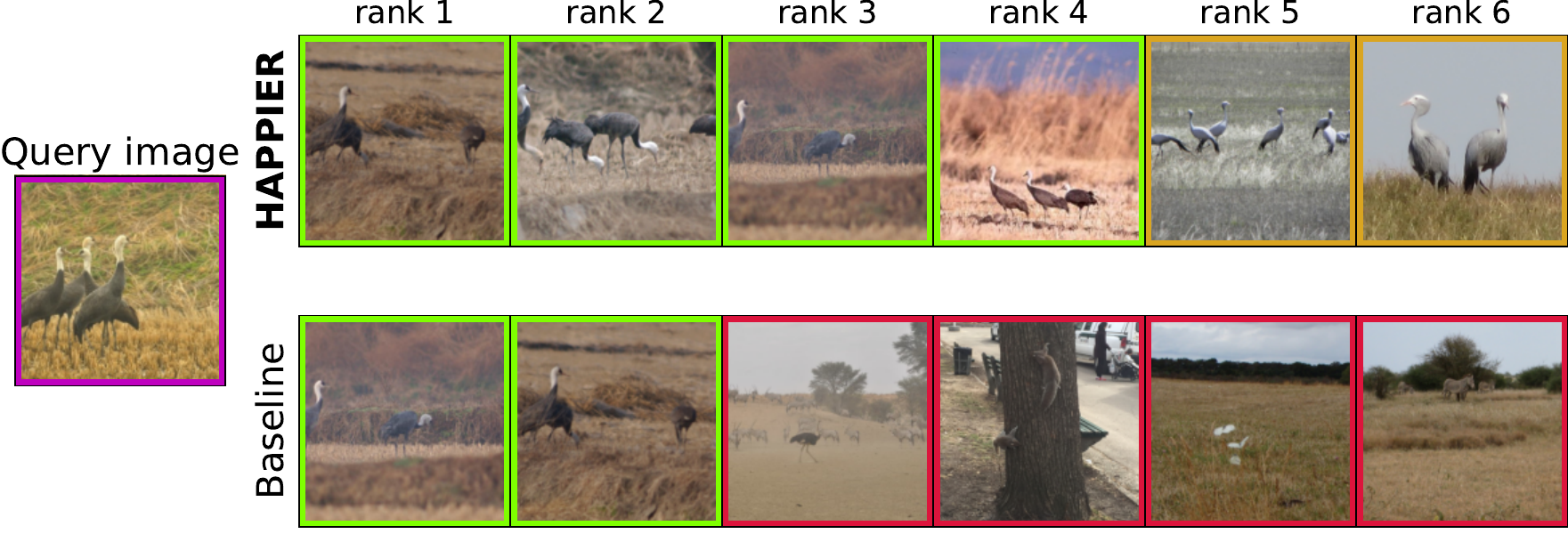}
        \caption{\textbf{HAPPIER} makes less severe mistakes. Its inversions are with instances sharing the same coarse label (in \textcolor{orange}{orange}) where the baseline (Smooth-AP~\cite{smoothap}) has inversion  with negative instances (in \textcolor{red}{red}).
        }
        \label{fig:sup_qual_inat_good}
    \end{subfigure}
    
    \begin{subfigure}[t]{\textwidth}
    \centering
    \includegraphics[width=0.95\textwidth]{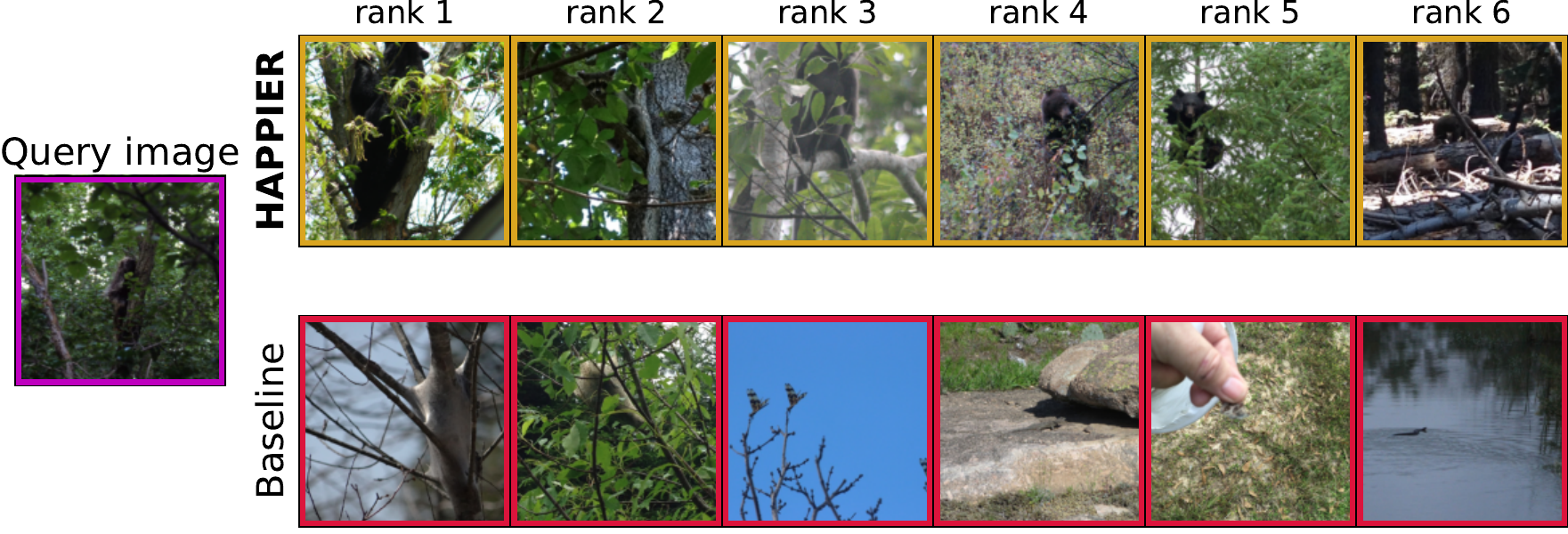}
    \caption{When models fail to retrieve the correct fine-grained images, \textbf{HAPPIER} still retrieves images with the same coarse label (in \textcolor{orange}{orange}) whereas the baseline (Smooth-AP~\cite{smoothap}) retrieves negative instances (in \textcolor{red}{red}).
    }
    \label{fig:sup_qual_inat_error}
    \end{subfigure}
\end{minipage}%
\vspace{-1.5em}
\end{figure*}

\subsubsection{Qualitative experiments}\label{sec:qual_results}

We assess qualitatively HAPPIER, including embedding space analysis and visualization of HAPPIER's retrievals.

\medbreak
\noindent\textbf{t-SNE: organization of the embedding space:}
In~\cref{fig:tsne_baseline,fig:tsne_happier}, we plot using t-SNE~\cite{tsne,chan2019gpu} how HAPPIER learns an embedding space on SOP ($L=2$) that is well-organized. We plot the mean vector of each fine-grained class and we assign the color based on the coarse level. We compare the t-SNE the embedding space of a baseline ( Smooth-AP~\cite{smoothap}) on~\cref{fig:tsne_baseline} and of HAPPIER in~\cref{fig:tsne_happier}. We cannot observe any clear clusters for the coarse level on~\cref{fig:tsne_baseline}, whereas we can appreciate the the quality of the hierarchical clusters formed on~\cref{fig:tsne_happier}.

\medbreak
\noindent\textbf{Controlled errors on iNat-base:}
Finally, we showcase in~\cref{fig:sup_qual_inat_good,fig:sup_qual_inat_error} errors of HAPPIER \vs a fine-grained baseline (Smooth-AP) on iNat-base. On~\cref{fig:sup_qual_inat_good}, we illustrate how a model trained with HAPPIER makes less severe mistakes than a model trained only on the fine-grained level. On~\cref{fig:sup_qual_inat_error}, we show an example where both models fail to retrieve the correct fine-grained instances, however the model trained with HAPPIER retrieves images that are semantically more similar to the query.
This shows the the robustness of HAPPIER's ranking.

% \clearpage
% \newpage
%\vspace{-1.2em}
\section{Conclusion}\label{sec:conclusion}

In this work we have introduced a general framework for rank losses optimization. It tackles two issues of rank losses optimization: 1) non-differentiability using smooth and upper bound rank approximation, 2) non-decomposability using an additional objective. We apply our framework to both fine-grained, by optimizing the AP and R@k, and hierarchical image retrieval, by optimizing the NDCG and the introduced $\hap$. We show that using our framework outperforms other rank loss surrogates on several standard fine-grained and hierarchical image retrieval benchmarks, including the hierarchical landmark dataset we introduce in this work. We also show that our framework sets state-of-the-art results for fine-grained image retrieval.

% or


%\newpage
%\clearpage
\vspace{-1em}

\ifCLASSOPTIONcompsoc
  \section*{Acknowledgments}
\else
  \section*{Acknowledgment}
\fi

\noindent This work was done under a grant from the the AHEAD ANR program (ANR-20-THIA-0002) and had access to HPC resources of IDRIS under the allocation AD011012645 made by GENCI.

\ifCLASSOPTIONcaptionsoff
  \newpage
\fi

\vspace{-1.5em}



\bibliographystyle{IEEEtran}
\bibliography{egbib}


% \end{thebibliography}


\vspace{-4em}
\begin{IEEEbiography}[{\includegraphics[height=0.9in,clip,keepaspectratio]{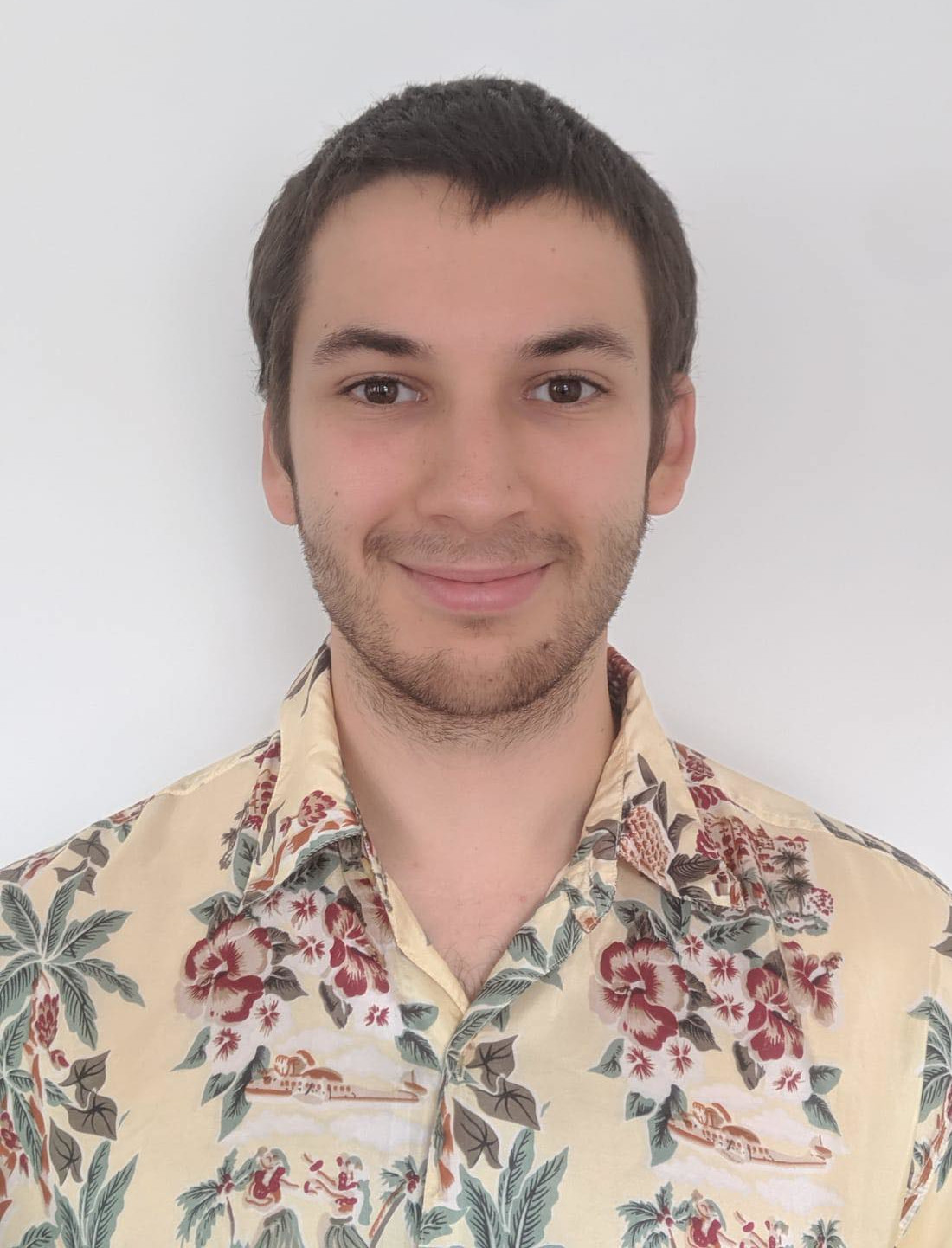}}]{Elias Ramzi} is a Ph.D. student in deep learning and computer vision at Cnam (Paris, France) and Coexya (Paris, France). He received a M.Eng. degree from CentraleSupélec in 2020. His research interests are deep learning for computer vision and image retrieval.
\end{IEEEbiography}
\vspace{-5em}
\begin{IEEEbiography}[{\includegraphics[height=0.9in,clip,keepaspectratio]{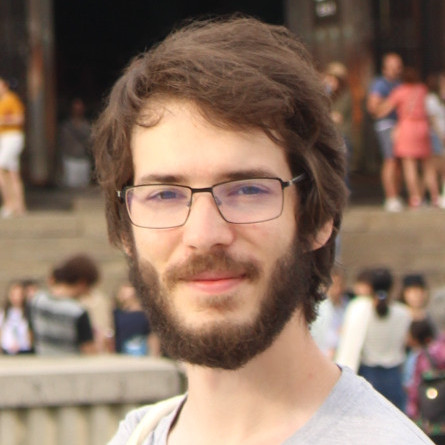}}]{Nicolas Audebert} is an associate professor of computer science at Cnam (Paris, France) since 2019. His research focus on deep representation learning for computer vision and Earth Observation. He graduated with a Ph.D. from the University of South Brittany in 2018 and a M.Eng. from Supélec in 2015.
\end{IEEEbiography}
\vspace{-5em}
\begin{IEEEbiography}[{\includegraphics[height=0.9in,clip,keepaspectratio]{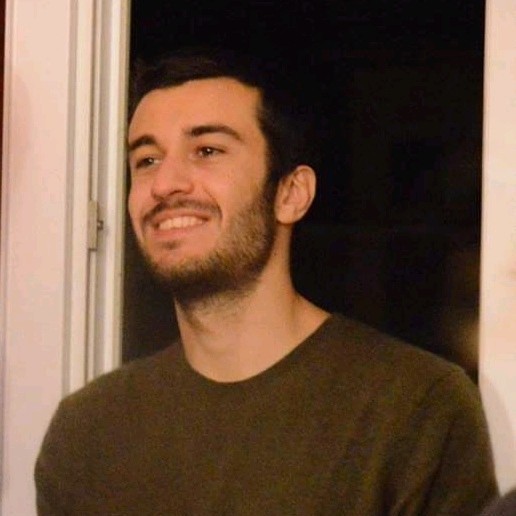}}]{Clément Rambour} is an Assistant Professor at the Cnam (Paris, France) since 2020. He obtained his MS degree from Sorbonne University in 2015 and completed his Ph.D. at Telecom Paris in 2019, specializing in 3D SAR imaging. His research primarily revolves around machine learning and deep learning for image understanding and signal processing. His work focus mainly on natural images as well as remote sensing and healthcare data.
\end{IEEEbiography}
\vspace{-5em}
\begin{IEEEbiography}[{\includegraphics[height=0.9in,clip,keepaspectratio]{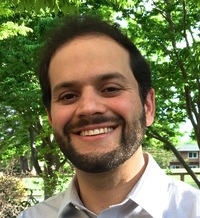}}]{André Araujo} is a Staff Software Engineer / Tech Lead Manager at Google Research. His current work focuses on computer vision and machine learning. He graduated with a Ph.D. from Stanford University in 2016.
\end{IEEEbiography}
\vspace{-5em}
\begin{IEEEbiography}[{\includegraphics[height=0.9in,clip,keepaspectratio]{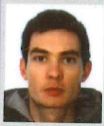}}]{Xavier Bitot} is a project leader at Coexya (France). He graduated with a M.Eng. from École Nationale Superieure de Physique de Strasbourg in 2002. He supervises the Coexya SIP (Paris) entity R\&D team on content based text and image classification and retrieval for industrial property, and is project manager of the Acsepto search software for Trademark and Industrial Design search software edited by Coexya.
\end{IEEEbiography}
\vspace{-5em}
\begin{IEEEbiography}[{\includegraphics[height=0.9in,clip,keepaspectratio]{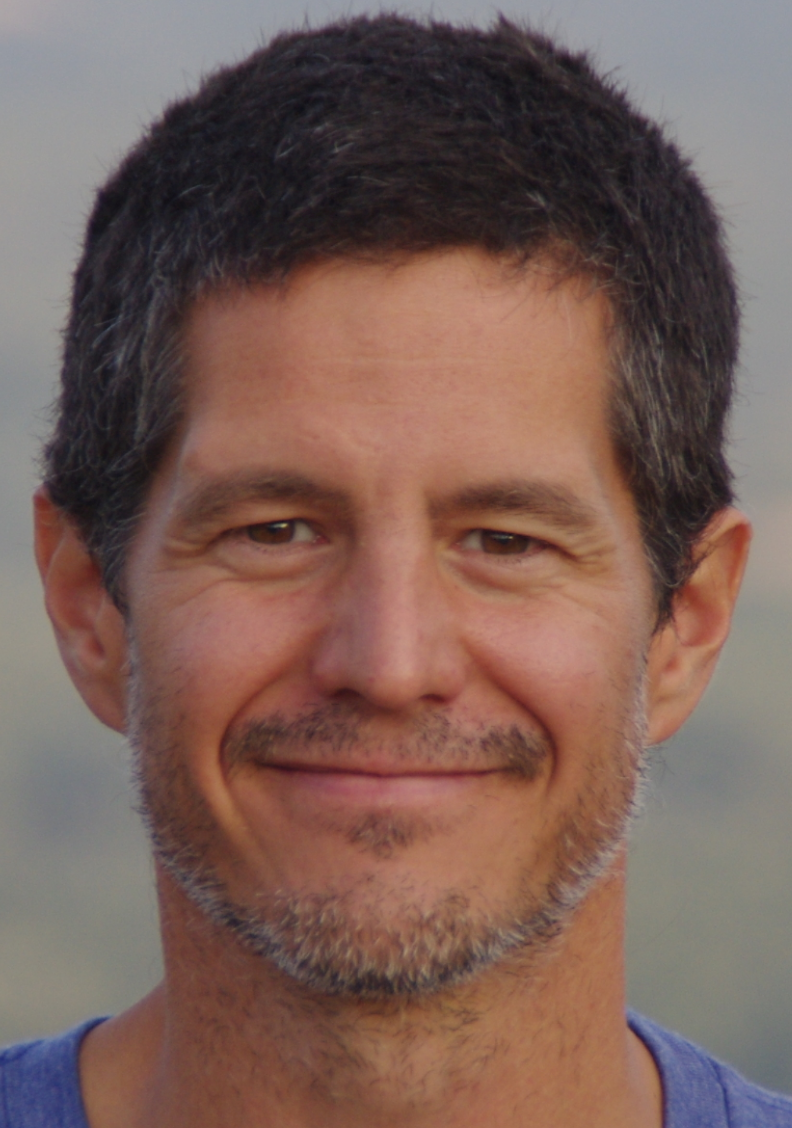}}]{Nicolas Thome} is a full professor at Sorbonne Université (Paris, France). His research interests include machine learning and deep learning for understanding low-level signals, e.g.. vision, time series, acoustics, etc. He also explores solutions for combining low-level and more higher-level data for multi-modal data processing. His current application domains are essentially targeted towards healthcare, autonomous driving and physics.
\end{IEEEbiography}

\newpage
\clearpage

\appendices

\section{Additional details on method}

\subsection{Decomposability gap: Average Precision}\label{sec:sup_dg}

We remind the reader of the definition of the decomposability gap given in~\cref{eq:decomposability-gap} of the main paper.

\begin{equation*}
    DG(\boldsymbol{\theta}) =  \frac{1}{K}  \sum_{b=1}^K \AP_i^b(\boldsymbol{\theta}) - \AP_i(\boldsymbol{\theta})
\end{equation*}

\begin{figure}[ht]
    \centering
    \includegraphics[width=0.5\textwidth]{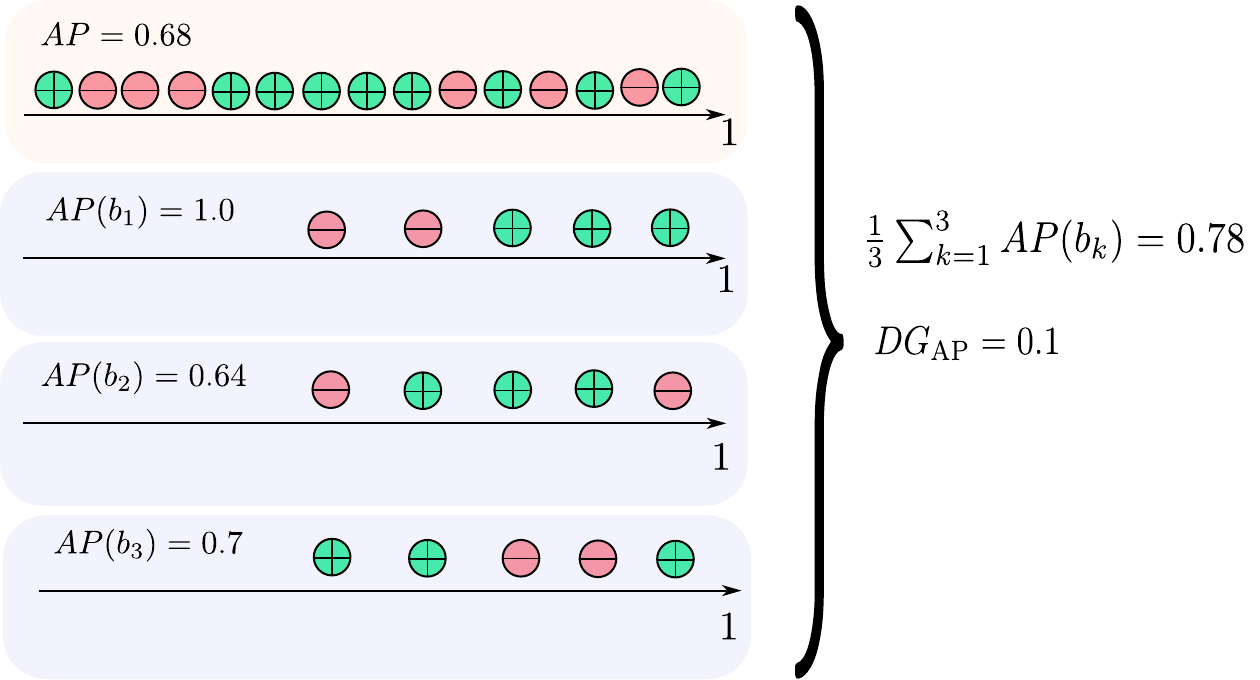}
    \caption{Illustration of the decomposability gap on a toy dataset.}
    \label{fig:dg_ap}
\end{figure}

We illustrates the decomposability gap, $DG$ with the toy dataset of \cref{fig:dg_ap}. The decomposability gap comes from the fact that the AP is not decomposable in mini-batches as we discuss in~\cref{sec:decomp}. The motivation behind $\Labs$ is thus to force the scores of the different batches to be calibrated between mini-batches.

\begin{table*}[t]
    \label{tab:compa_bounds} 
    \centering
    \begin{tabularx}\textwidth{@{}lML@{}}
        \toprule
         without $\Labs$ & 
         0 \leq DG \leq 1 - \frac{1}{\sum_{b=1}^K |\Omega^+_b|}\left( \sum_{b=1}^K \sum_{j=1}^{|\Omega^+_b|} \frac{j + |\Omega^+_1| + \dots + |\Omega^+_{b-1}|}{j + |\Omega^+_1| + \dots + |\Omega^+_{b-1}| + |\Omega^-_1| + \dots + |\Omega^-_{b-1}|} \right ) 
         & eq:upperBound
         \\
& \\
        with $\Labs$ & 
        0 \leq DG \leq  1 - \frac{1}{\sum_{b=1}^K |\Omega^+_b|} \Bigg( \sum_{b=1}^K \bigg[ \sum_{j=1}^{G^+_b} \frac{j + G^+_1 + \dots + G^+_{b-1}}{j + G^+_1 + \dots + G^+_{b-1} + E^-_1 + \dots E^-_{b-1}} + \sum_{j=1}^{E^+_b} \frac{j + G^+_{b} + |\Omega^+_1| + \dots + |\Omega^+_{b-1}|}{j + G^+_{b} + |\Omega^+_1| + \dots + |\Omega^+_{b-1}| + |\Omega^-_1| + \dots + |\Omega^-_{b-1}|} \bigg]  \Bigg) 
        & eq:upperBoundRefined
        \\
         \bottomrule
    \end{tabularx}
    \vspace{-1.5em}
\end{table*}

\subsection{Upper bound on the decomposabilty gap}\label{sec:sup_upperbound}

To formalize this idea, we provide a theoretical analysis of the impact on the global ranking of $\Labs$ in \cref{eq:labs} for standard image retrieval. 
Firstly, we can see that if $\Labs=0$ on every batch, the overall $\mathcal{M}$ and $\mathcal{M}_b$ is maximal (1), \ie $DG(\boldsymbol{\theta}) = 0$ and we get a decomposable $\mathcal{M}$. In a more general setting, we show that minimizing $\Labs$ on each batch reduces the decomposability gap, hence improving the decomposability of the $\mathcal{M}$.

Let us consider $K$ batches $\{\mathcal{B}_b\}_{b\in\{1:K\}}$ of batch size $B$ divided in $\Omega^+_{b}$ positive instances and $\Omega^-_{b}$ negative instances w.r.t. the query $\boldsymbol{q_i}$.
To give some insight we assume that $ \mathcal{M}_b = 1$. This results in the upper bound of $DG$ given in~\cref{eq:upperBound}. This upper bound of the decomposability gap is given in the worst case for the global $\mathcal{M}$: the global ranking is built from the juxtaposition of the batches {(see proof bellow)}.

We can tighten this upper bound by introducing the decomposability loss $\Labs$ and constraining the scores of positive and negative instances to be well calibrated. On each batch we define the following quantities $E^-_b = \sum_{j \in \Omega^-_b} \mathds{1}(s_j > \beta)$ which are the number of negative instances that do not respect the constraints and $G^-_b = \sum_{j \in \Omega^-_b} \mathds{1}(s_j \leq \beta)$ the number of negative instances that do. We similarly define $E^+_b$ and $G^+_b$.
Giving the upper bound of~\cref{eq:upperBoundRefined}. $\Labs$ loss directly optimizes this upper bound (by explicitly optimizing $E^-_b, E^+_b, E^+_{b}, G^+_{b}$), making it tighter, hence improving the decomposability of $\mathcal{M}$.

\paragraph*{Proof of~\cref{eq:upperBound}: Upper bound on the $DG$ with no $\Labs$} We choose a setting for the proof of the upper bound similar to the one used for training, \ie all the batch have the same size, and the number of positive instances per batch (\ie $\mathcal{P}_i^b$) is the same.

\begin{figure}[ht]
     \centering
     \includegraphics[width=0.3\textwidth]{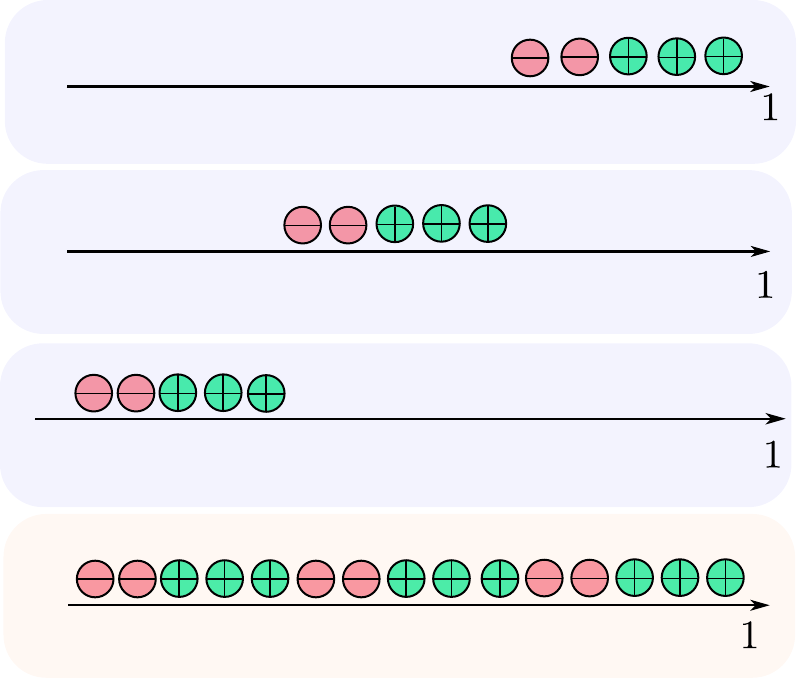}
     \caption{The worst case when computing the global AP would be that each batch is juxtaposed.}
     \label{fig:worst_case_global_ap}
 \end{figure}

\cref{eq:upperBound} gives an upper bound in the worst case: when the AP has the lowest value guaranteed by the AP on each batch. We illustrate this case in~\cref{fig:worst_case_global_ap}.

In \cref{eq:upperBound} the $1$ in the right hand term comes from the average of AP over all batches:
\begin{equation*}
    \frac{1}{K}\sum_{b=1}^K AP_i^b(\theta)=1
\end{equation*}

We then justify the term in the parenthesis of~\cref{eq:upperBound}, which is the lower bound of the AP.
In the global ordering the positive instances are ranked after all the positive instances from previous batches giving the following $\rank^+$: ${j+|\mathcal{P}_i^1|+\dots+|\mathcal{P}_i^{b-1}|}$, with $j$ the $\rank^+$ in the batch, positive instances are also ranked after all negative instances from previous batches giving $\rank^-$: ${|\mathcal{N}_i^1|+\dots+|\mathcal{N}_i^{b-1}|}$. Therefore we obtain the resulting upper bound of~\cref{eq:upperBound}.

%\medbreak
\paragraph*{Proof of~\cref{eq:upperBoundRefined}: Upper bound on the $DG$ with $\Labs$} We refine the upper bound on $DG$ in~\cref{eq:upperBoundRefined} by adding $\Labs$ which calibrates the absolute scores across the mini-batches.

We now write that each positive instance that respects the constraint of $\Labs$ is ranked after the positive instances of previous batch that respect the constraint giving the following $\rank^+$: ${j+G_1^++\dots+G_{b-1}^+}$, with $j$ the $\rank^+$ in the current batch. Positive instances are also ranked after the negative instances of previous batches that do not respect the constraints yielding $\rank^-$ : ${E_1^-+\dots+E_{b-1}^-}$.

We then write that positive instances that do not respect the constraints are ranked after all positive instances from previous batches and the positive instances respecting the constraints of the current batch giving $\rank^+$ : ${j+G_b^+|\mathcal{P}_i^1|+\dots+|\mathcal{P}_i^{b-1}|}$. They also are ranked after all the negative instances from previous batches giving $\rank^-$ : ${|\mathcal{N}_i^1|+\dots+|\mathcal{N}_i^{b-1}|}$. Resulting in~\cref{eq:upperBoundRefined}.

\subsection{Choice of $\delta$}\label{sec:choice_of_delta}

\begin{figure}[ht]
    \centering
    \includegraphics[scale=.5]{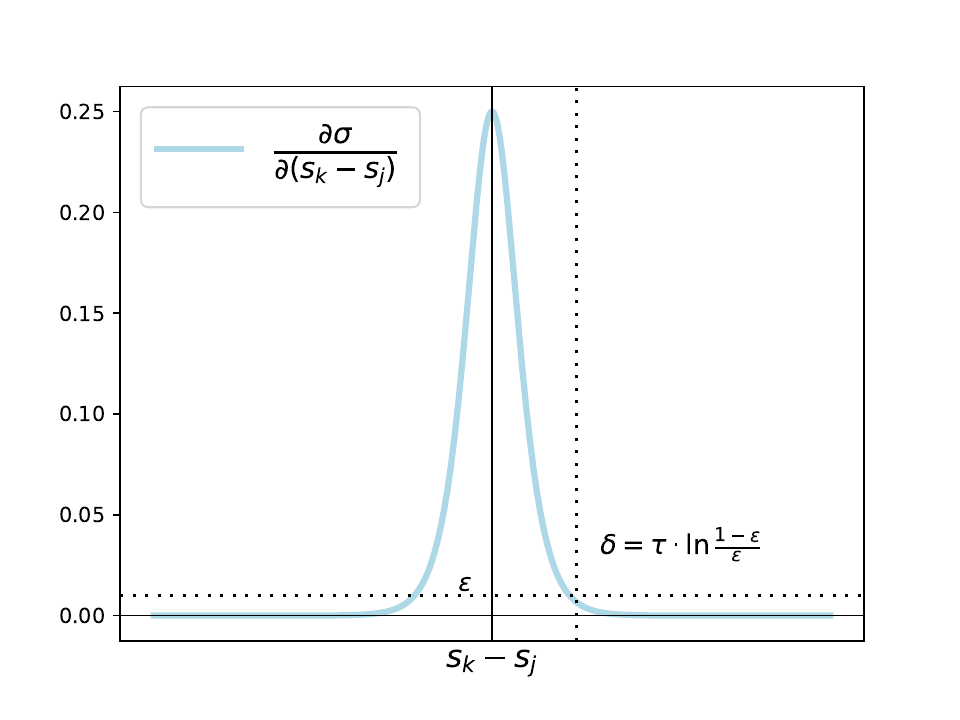}
    \caption{Gradient of the temperature scaled sigmoid ($\tau=0.01$) \vs the difference of scores $s_k-s_j$ of a negative pair.}
    \label{fig:choice_of_delta}
\end{figure}

In the main paper we introduce $\delta$ in~\cref{eq:h_minus} to define $H^-$. We choose $\delta$ as the point where the gradient of the sigmoid function becomes low $< \epsilon$, and we then have $\delta=\tau\cdot\ln\frac{1-\epsilon}{\epsilon}$. This is illustrated in \cref{fig:choice_of_delta}. For our experiments we use $\epsilon=10^{-2}$ giving $\delta\simeq0.05$.

\clearpage
\section{Additional details on $\hap$}

\subsection{Details on $\hrank$}\label{sec:sup_hrank}

We define the $\hrank$ in the main paper as:

\begin{equation}
    \hrank(k) = \rel(k) + \sum_{j\in\Omega^+} \min(\rel(k), \rel(j))\cdot H(s_j-s_k) ~.
    \label{eq:sup_hierarchical_rank}
\end{equation}

We detail in~\cref{fig:supp_hrank_figure} how the $\hrank$ in~\cref{eq:sup_hierarchical_rank} is computed in the example from~\cref{fig:hierarchical_tree} of the main paper. Given a ``Lada \#2'' query, we set the relevances as follows: if $k\in\Omega^{(3)}$ (\ie $k$ is also a ``Lada \#2''), $\rel(k)=1$; if $k\in\Omega^{(2)}$ (\ie $k$ is another model of ``Lada''), $\rel(k)=2/3$; and if $k\in\Omega^{(1)}$ ($k$ is a ``Car''), $\rel(k)=1/3$. Relevance of negatives (other vehicles) is set to 0.

\definecolor{amethyst}{rgb}{0.6, 0.4, 0.8}
\begin{figure*}
    \centering
    \includegraphics[width=0.9\textwidth]{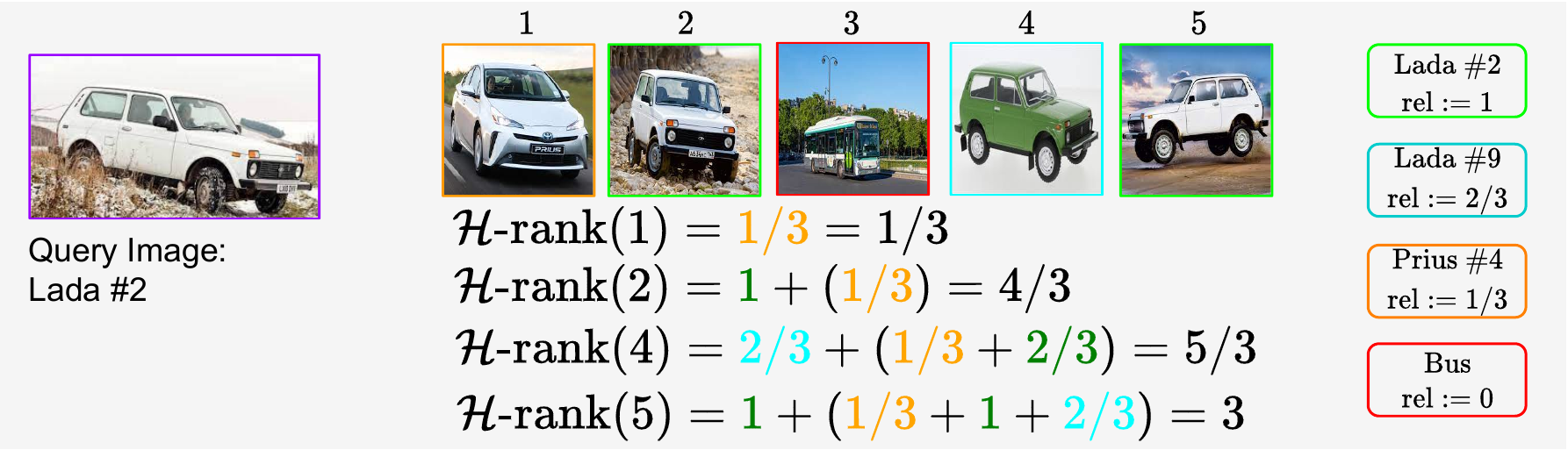}
    \caption{$\hrank$ for each retrieval results given a ``\textcolor{Green}{Lada \#2}'' \textcolor{amethyst}{query} with relevances of~\cref{sec:sup_hrank} and the hierarchical tree of~\cref{fig:hierarchical_tree} in the main paper.}
    \label{fig:supp_hrank_figure}
\end{figure*}

In this instance, $\hrank(2)=4/3$ because $\rel(2)=1$ and $\min(\rel(1), \rel(2)) = \rel(1) = 1/3$. Here, the closest common ancestor in the hierarchical tree shared by the query and instances $1$ and $2$ is ``Cars''. For binary labels, we would have $\rank^+(2)=1$; this would not take into account the semantic similarity between the query and instance $1$.

\subsection{Details on $\hap$}\label{sec:sup_detail_hap}

We define $\hap$ in the main paper as:

\begin{equation}\label{eq:sup_def_hap}
    \hap = \frac{1}{\sum_{k\in\Omega^+}\rel(k)} \sum_{k\in\Omega^+} \frac{\hrank(k)}{\rank(k)}
\end{equation}

\begin{figure*}[t]
    \centering
    \includegraphics[width=0.9\textwidth]{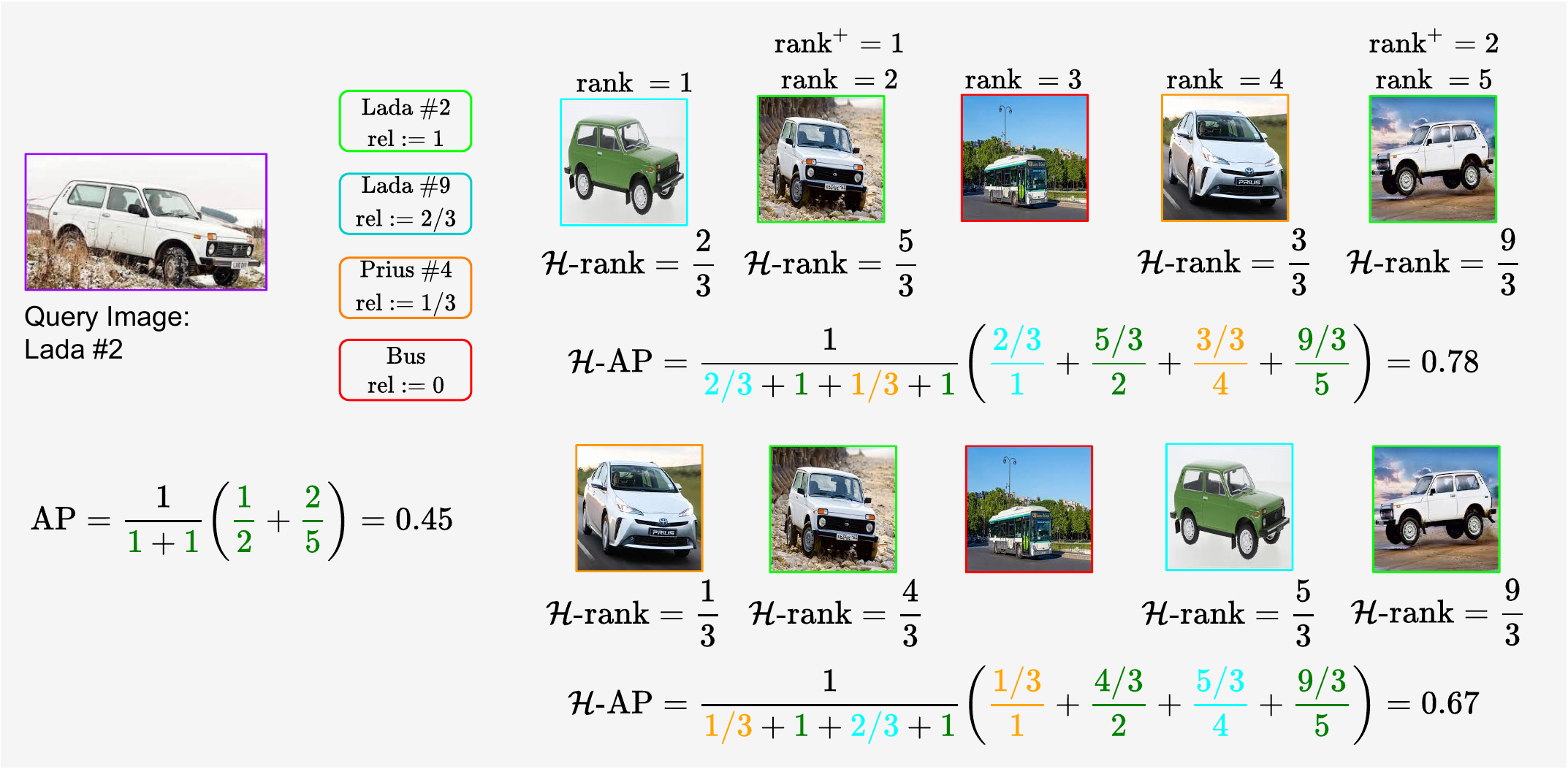}
    \caption{AP and $\hap$ for two different rankings when Given a ``\textcolor{Green}{Lada \#2}'' \textcolor{amethyst}{query} and relevances of~\cref{sec:sup_hrank}. The $\hap$ of the top row is greater (0.78) than the bottom one's (0.67) as the error in $\rank=1$ is less severe for the top row. Whereas the AP is the same for both rankings (0.45).}
    \label{fig:sup_dif_ap_hap}
\end{figure*}

We illustrate in~\cref{fig:sup_dif_ap_hap} how the $\hap$ is computed for two rankings. We use the same relevances as in~\cref{sec:sup_hrank}. The $\hap$ of the first example is greater ($0.78$) than of the second one ($0.67$) because the error is less severe. On the contrary, the AP only considers binary labels and is the same for both rankings ($0.45$).

\medbreak

\textcolor{black}{One property of AP is that it can be interpreted as the area under the precision-recall curve. $\hap$ from~\cref{eq:sup_def_hap} can also be interpreted as the area under a hierarchical-precision-recall curve by defining a Hierarchical Recall ($\mathcal{H}\text{-R@k}$) and a Hierarchical Precision ($\mathcal{H}\text{-P@k}$) as:}

\begin{align}
    &\mathcal{H}\text{-R@k} = \frac{\sum_{j=1}^k \rel(j)}{\sum_{j\in\Omega^+} \rel(j)} \label{eq:hierarchical_recall}\\
    &\mathcal{H}\text{-P@k} = \frac{\sum_{j=1}^k \min(\rel(j),\rel(k))}{k\cdot\rel(k)} \label{eq:hierarchical_precision}
\end{align}

\textcolor{black}{So that $\hap$ can be re-written as:}
\begin{equation}
\label{eq:hap_generalizes_ap}
    \hap = \sum_{k=1}^{|\Omega|} (\mathcal{H}\text{-R@k}-\mathcal{H}\text{-R@k-1}) \times \mathcal{H}\text{-P@k}
\end{equation}

\textcolor{black}{\cref{eq:hap_generalizes_ap} recovers~\cref{eq:def_hap} from the main paper, meaning that $\hap$ generalizes this property of AP beyond binary labels. To further motivate $\hap$ we will justify the normalization constant for $\hap$, and show that $\hap$, $\mathcal{H}\text{-R@k}$ and $\mathcal{H}\text{-P@k}$ are consistent generalization of AP, R@k, P@k.}

\subsubsection{Normalization constant for $\hap$}\label{sec:sup_normalization_constant_hap}

When all instances are perfectly ranked, all instances $j$ that are ranked before instance $k$ ($s_j\geq s_k$) have a relevance that is higher or equal than $k$'s, \ie $\rel(j)\geq\rel(k)$ and \\ ${\min(\rel(j),\rel(k))=\rel(k)}$. So, for each instance $k$:

\begin{align*}
    \hrank(k) &= \rel(k) + \sum_{j\in\Omega^+} \min(\rel(k), \rel(j))\cdot H(s_j-s_k) \\
    &= \rel(k) + \sum_{j\in\Omega^+} \rel(k)\cdot H(s_j-s_k) \\
    &= \rel(k) \cdot \left( 1 + \sum_{j\in\Omega^+} H(s_j-s_k) \right) = \rel(k)\cdot\rank(k)
\end{align*}
The total sum $\sum_{k\in\Omega^+} \frac{\hrank(k)}{\rank(k)} = \sum_{k\in\Omega^+} \rel(k)$. This means that we need to normalize by $\sum_{k\in\Omega^+} \rel(k)$ in order to constrain $\hap$ between 0 and 1. This results in the definition of $\hap$ from~\cref{eq:sup_def_hap}.

\subsubsection{$\hap$ is a consistent generalization of AP}\label{sec:sup_hap_consistent_ap} In a binary setting, AP is defined as follows:

\begin{equation}
    \AP = \frac{1}{|\Omega^+|} \sum_{k\in\Omega^+} \frac{\rank^+(k)}{\rank(k)}
\end{equation}

$\hap$ is equivalent to AP in a binary setting ($L=1$). Indeed, the relevance function is $1$ for fine-grained instances and 0 otherwise in the binary case. Therefore $\hrank(k) = 1 + \sum_{j\in\Omega^+} H(s_j-s_k)$ which is the same definition as $\rank^+$ in AP. Furthermore the normalization constant of $\hap$, $\sum_{k\in\Omega^+} \rel(k)$, is equal to the number of fine-grained instances in the binary setting, \ie $|\Omega^+|$. This means that $\hap=\AP$ in this case.

\medbreak
$\mathcal{H}\text{-R@k}$ is also a consistent generalization of R@k, indeed:

\begin{align*}
    \mathcal{H}\text{-R@k} &= \frac{\sum_{j=1}^k \rel(j)}{\sum_{j\in\Omega^+} \rel(j)} = \frac{\sum_{j=1}^k \mathds{1}(k\in\Omega^+)}{\sum_{j\in\Omega^+} \mathds{1}(k\in\Omega^+)} \\
    &= \frac{\text{\# number of positive before k}}{|\Omega^+|} \\
    &= R@k
\end{align*}

\medbreak
Finally, $\mathcal{H}\text{-P@k}$ is also a consistent generalization of P@k:

\begin{align*}
    \mathcal{H}\text{-P@k} &= \frac{\sum_{j=1}^k \min(\rel(j),\rel(k))}{k\cdot\rel(k)} \\
    &= \frac{\text{\# number of positive before k}}{k} \\
    &= P@k
\end{align*}

\subsubsection{Link between $\hap$ and the weighted average of AP}\label{sec:hap_and_weighted_ap}

Let us define the AP for the semantic level $l\geq1$ as the binary AP with the set of positives being all instances that belong the same level, \ie $\Omega^{+,l} = \bigcup_{q=l}^L \Omega^{(q)}$:

\begin{equation}\label{eq:sup_ap_level}
    \AP^{(l)} = \frac{1}{|\Omega^{+,l}|} \sum_{k\in\Omega^{+,l}} \frac{\rank^{+,l}(k)}{\rank(k)}, \; \rank^{+,l}(k) = 1 + \sum_{j\in\Omega^{+,l}} H(s_j-s_k)
\end{equation}

\fbox{
\parbox{0.9\linewidth}{%
\begin{property}\label{prop:link_hap_ap} For any relevance function  $\rel(k) = \sum_{p=1}^l \frac{w_p}{|\Omega^{+,q}|}, \, k\in\Omega^{(l)}$, with positive weights $\{w_l\}_{l \in \llbracket 1;L\rrbracket}$ such that $\sum_{l=1}^L w_l =1$:

\begin{equation*}
    \hap = \sum_{l=1}^L w_l \cdot AP^{(l)}
\end{equation*}

\ie $\hap$ is equal the weighted average of the AP at all semantic levels.
\end{property}
}
}

\medbreak
\medbreak
\textbf{Proof of Property~\ref{prop:link_hap_ap}}

\medbreak
Denoting $\Sigma w\AP:=\sum_{l=1}^L w_l \cdot AP^{(l)}$, we obtain from~\cref{eq:sup_ap_level}:

\begin{equation}
    \Sigma w\AP = \sum_{l=1}^L  w_l \cdot \frac{1}{|\Omega^{+,l}|} \sum_{k\in\Omega^{+,l}} \frac{\rank^{+,l}(k)}{\rank(k)}
\end{equation}

We define $\hat{w}_l = \frac{w_l}{|\Omega^{+,l}|}$ to ease notations, so:

\begin{equation}
    \Sigma w\AP = \sum_{l=1}^L  \hat{w}_l \sum_{k\in\Omega^{+,l}} \frac{\rank^{+,l}(k)}{\rank(k)} 
\end{equation}

We define $\mathds{1}(k,l) = \mathds{1}\left[k\in\Omega^{+,l} \right]$ so that we can sum over $\Omega^+$ instead of $\Omega^{+,l}$ and inverse the summations. Note that rank does not depend on $l$, on contrary to $\rank^{+,l}$.
\begin{align}
    \Sigma w\AP &= \sum_{l=1}^L \sum_{k\in\Omega^+} \frac{\hat{w}_l \cdot \mathds{1}(k,l) \cdot \rank^{+,l}(k)}{\rank(k)} \\
    &= \sum_{k\in\Omega^+} \sum_{l=1}^L \frac{\hat{w}_l \cdot \mathds{1}(k,l) \cdot \rank^{+,l}(k)}{\rank(k)} \\
    &= \sum_{k\in\Omega^+} \frac{\sum_{l=1}^L  \mathds{1}(k,l) \cdot \hat{w}_l \cdot \rank^{+,l}(k)}{\rank(k)} \label{eq:it_is_inverted}
\end{align}

We replace $\rank^{+,l}$ in~\cref{eq:it_is_inverted} with its definition from~\cref{eq:sup_ap_level}:
\begin{align}
    \Sigma w\AP &= \sum_{k\in\Omega^+} \frac{\sum_{l=1}^L  \mathds{1}(k,l) \cdot \hat{w}_l \cdot \left(1 + \sum_{j\in\Omega^{+,l}} H(s_j-s_k)\right)}{\rank(k)} \\
    &= \sum_{k\in\Omega^+} \frac{\sum_{l=1}^L  \mathds{1}(k,l) \cdot \hat{w}_l + \sum_{l=1}^L \sum_{j\in\Omega^{+,l}} \mathds{1}(k,l) \cdot \hat{w}_l \cdot H(s_j-s_k)}{\rank(k)} \\
    &= \sum_{k\in\Omega^+} \frac{\sum_{l=1}^L  \mathds{1}(k,l) \cdot \hat{w}_l +  \sum_{l=1}^L \sum_{j\in\Omega^+} \mathds{1}(j,l) \cdot \mathds{1}(k,l) \cdot \hat{w}_l \cdot H(s_j-s_k)}{\rank(k)} \\
    &= \sum_{k\in\Omega^+} \frac{\sum_{l=1}^L  \mathds{1}(k,l) \cdot \hat{w}_l + \sum_{j\in\Omega^+} \sum_{l=1}^L \mathds{1}(j,l) \cdot \mathds{1}(k,l) \cdot \hat{w}_l \cdot H(s_j-s_k)}{\rank(k)} \label{eq:nearl_hrank}
\end{align}

We define the following relevance function:
\begin{equation}\label{eq:temporary_relevance}
    \rel(k)=\sum_{l=1}^L  \mathds{1}(k,l) \cdot \hat{w}_l
\end{equation}
By construction of $\mathds{1}(\cdot,l)$: 
\begin{equation}\label{eq:see_the_min}
    \sum_{l=1}^L \mathds{1}(j,l) \cdot \mathds{1}(k,l) \cdot \hat{w}_l = \min(\rel(k), \rel(j))
\end{equation}

Using the definition of the relevance function from~\cref{eq:temporary_relevance} and~\cref{eq:see_the_min}, we can rewrite~\cref{eq:nearl_hrank} with $\hrank$:

\begin{align}
    \Sigma w\AP &= \sum_{k\in\Omega^+} \frac{\rel(k) + \sum_{j\in\Omega^+} \min(\rel(j), \rel(k)) \cdot H(s_j-s_k)}{\rank(k)} \\
    &= \sum_{k\in\Omega^+} \frac{\hrank(k)}{\rank(k)} \label{eq:almost_hap}
\end{align}

\cref{eq:almost_hap} lacks the normalization constant $\sum_{k\in\Omega^+} \rel(k)$ in order to have the same shape as $\hap$ in~\cref{eq:sup_def_hap}. So we must prove that $\sum_{k\in\Omega^+} \rel(k) = 1$:
\begin{align}
    \sum_{k\in\Omega^+} \rel(k) &= \sum_{k\in\Omega^+} \sum_{l=1}^L  \mathds{1}(k,l) \cdot \hat{w}_l \\
    &= \sum_{l=1}^L |\Omega^{(l)}| \sum_{p=1}^l \hat{w}_p \\
    &= \sum_{l=1}^L |\Omega^{(l)}| \sum_{p=1}^l \frac{w_p}{|\Omega^{+,p}|} \\
    &= \sum_{l=1}^L |\Omega^{(l)}| \sum_{p=1}^l \frac{w_p}{|\bigcup_{q=p}^L \Omega^{(q)}|} \\
    &= \sum_{l=1}^L |\Omega^{(l)}| \sum_{p=1}^l \frac{w_p}{\sum_{q=p}^L |\Omega^{(q)}|} \\
    &= \sum_{l=1}^L  \sum_{p=1}^l \frac{|\Omega^{(l)}| \cdot w_p}{\sum_{q=p}^L |\Omega^{(q)}|} \\
    &= \sum_{p=1}^L  \sum_{l=p}^L \frac{|\Omega^{(l)}| \cdot w_p}{\sum_{q=p}^L |\Omega^{(q)}|} \\
    &= \sum_{p=1}^L w_p \cdot \frac{ \sum_{l=p}^L |\Omega^{(l)}| }{\sum_{q=p}^L |\Omega^{(q)}|} \\
    &= \sum_{p=1}^L w_p = 1
\end{align}

We have proved that $\Sigma w\AP = \hap$ with the relevance function of~\cref{eq:temporary_relevance}:
\begin{equation}
    \Sigma w\AP = \frac{1}{\sum_{k\in\Omega^+}\rel(k)} \sum_{k\in\Omega^+} \frac{\hrank(k)}{\rank(k)} = \hap
\end{equation}

Finally we show, for an instance $k\in\Omega^{(l)}$, :

\begin{equation}
    \rel(k)=\sum_{p=1}^L  \mathds{1}(k,p) \cdot \hat{w}_p = \sum_{p=1}^l \cdot \hat{w}_p = \sum_{p=1}^l \frac{w_p}{|\Omega^{+,p}|}
\end{equation}
\ie the relevance of~\cref{eq:temporary_relevance} is the same as the relevance of Property~\ref{prop:link_hap_ap}. This concludes the proof of Property~\ref{prop:link_hap_ap}. $\square$

\clearpage
\section{Additional experimental results}

\subsection{Additional hierarchical results}

\subsubsection{ASI.}\label{sec:sup_asi}

The ASI~\cite{fagin2003comparing} measures at each rank $n\leq N$ the set intersection proportion ($SI$) between the ranked list $a_1,\dots,a_N$ and the ground truth ranking $b_1,\dots,b_N$, with $N$ the total number of positives. As it compares intersection the ASI can naturally take into account the different levels of semantic: 

\begin{align*}
    SI(n) &= \frac{|\{a_1,\dots,a_n\}\cap\{b_1,\dots,b_n\}|}{n} \\
    ASI &= \frac{1}{N} \sum_{n=1}^N SI(n) 
\end{align*}

\subsubsection{Dynamic metric learning results.}
On~\cref{tab:main_dyml}, we evaluate HAPPIER on the recent DyML benchmarks~\cite{sun2021dynamic}. HAPPIER again shows significant gains in mAP and ASI compared to methods only trained on fine-grained labels, \eg +9pt in mAP and +10pt in ASI on DyML-V. HAPPIER also outperforms other hierarchical baselines: +4.8pt mAP on DyML-V, +0.9 on DyML-A and +1.8 on DyML-P. In R@1, HAPPIER performs on par with other methods on DyML-V and outperforms other hierarchical baselines by a large margin on DyML-P: 63.7 \vs 60.8 for $\Sigma$NSM. Interestingly, HAPPIER also consistently outperforms CSL~\cite{sun2021dynamic} on its own datasets\footnote{CSL's score  on~\cref{tab:main_dyml} are above those reported in~\cite{sun2021dynamic}; personal discussions with the authors~\cite{sun2021dynamic} validate that our results are valid for CSL.}.

\begin{table*}[ht]
    % \vspace{-0.6\intextsep}
    % \setlength\tabcolsep{0.3pt}
    \caption{Performance comparison on Dynamic Metric Learning benchmarks~\cite{sun2021dynamic}.
    }
    \label{tab:main_dyml} 
    \centering
    \begin{tabularx}{\textwidth}{l l YYY | YYY | YYY }
        \toprule
         & \multirow{2}{*}{Method} & \multicolumn{3}{c|}{DyML-Vehicle} & \multicolumn{3}{c|}{DyML-Animal} & \multicolumn{3}{c}{DyML-Product}\\
        \cmidrule{3-11}
         && mAP & ASI & R@1 & mAP & ASI & R@1 & mAP & ASI & R@1 \\
         \midrule
         \multirow{4}{*}{\rotatebox[origin=c]{90}{Fine}}
         & $\text{TL}_{\text{SH}}$~\cite{wu2017sampling} & 26.1 & 38.6 & 84.0 & 37.5 & 46.3 & 66.3 & 36.32 & 46.1 & 59.6 \\
         & NSM~\cite{norm_softmax} & 27.7 & 40.3 & 88.7 & 38.8 & 48.4 & \underline{69.6} & 35.6 & 46.0 & 57.4 \\
         & \textcolor{black}{Smooth-AP~\cite{smoothap}} & 27.1 & 39.5 & 83.8 & 37.7 & 45.4 & 63.6 & 36.1 & 45.5 & 55.0  \\
         & ROADMAP~\cite{ramzi2021robust} & 27.1 & 39.6 & 84.5 & 34.4 & 42.6 & 62.8 & 34.6 & 44.6 & \underline{62.5} \\
        \midrule
        \multirow{6}{*}{\rotatebox[origin=c]{90}{Hier.}}
         & $\Sigma\text{TL}_{\text{SH}}$~\cite{wu2017sampling} & 25.5 & 38.1 & 81.0 & 38.9 & 47.2 & 65.9 & \underline{36.9} & 46.3 & 58.5 \\
         & $\Sigma$NSM~\cite{norm_softmax} & \underline{32.0} & \underline{45.7} & \textbf{89.4} & \underline{42.6} & \underline{50.6} & \textbf{70.0} & 36.8 & \underline{46.9} & 60.8 \\
        & CSL~\cite{sun2021dynamic} & 30.0 & 43.6 & 87.1 & 40.8 & 46.3 & 60.9 & 31.1 & 40.7 & 52.7 \\
        & CLCD-ACR~\cite{zheng2022dynamic} & 16.0 & 42.9 &-& 36.0 & 57.1 &-& 29.4 & 58.8 & - \\
        & CLCD-ICR~\cite{zheng2022dynamic} & 16.6 & 43.7 &-& 35.7 & 56.0 &-& 30.2 & 59.5 & - \\
         \cmidrule{2-11}
        & \textbf{HAPPIER} & \textbf{37.0} & \textbf{49.8} & \underline{89.1} &  \textbf{43.8} & \textbf{50.8} & 68.9 & \textbf{38.0} & \textbf{47.9} & \textbf{63.7}\\
         \bottomrule
    \end{tabularx}
    %  \vspace {-13pt}
\end{table*}

\subsubsection{Detailed evaluation}\label{sec:sup_detailed_eval}

\cref{tab:detail_sop_inat_base} shows the different methods' performances on all semantic hierarchy levels. We evaluate HAPPIER and $\text{HAPPIER}_{\text{F}}$ with $\alpha=5$ on SOP and $\alpha=3$ on iNat-base. Similarly to~\cref{tab:main_detail_inat_full}, \cref{tab:detail_sop_inat_base} shows that HAPPIER gives the best performances at the coarse level, with a significant boost compared to fine-grained methods, \eg +43.9pt AP compared to the best non-hierarchical $\text{TL}_{\text{SH}}$~\cite{wu2017sampling} on SOP. HAPPIER even outperforms the best fine-grained methods in R@1 on iNat-base, but is slightly below on SOP. $\text{HAPPIER}_{\text{F}}$ performs on par with the best methods at the finest level on SOP, while further improving performances on iNat-base, and still significantly outperforms fine-grained methods at the coarse level.

\begin{table*}[ht]
    % \vspace{-0.4\intextsep}
    \setlength\tabcolsep{2pt}
    \caption{Comparison of HAPPIER \vs methods trained only on fine-grained labels on SOP and iNat-base. Metrics are reported for both semantic levels.}
    \label{tab:detail_sop_inat_base} 
    \centering
    \begin{tabularx}{\textwidth}{ l l YYY | YYY}
        \toprule
         & & \multicolumn{3}{c|}{SOP} & \multicolumn{3}{c}{iNat-base} \\
        \cmidrule{3-8}
         & & \multicolumn{2}{c}{Fine} & \multicolumn{1}{c|}{Coarse} & \multicolumn{2}{c}{Fine} & \multicolumn{1}{c}{Coarse}  \\
         & Method & R@1 & AP & AP & R@1 & AP & AP \\
         \midrule
         \multirow{5}{*}{\rotatebox[origin=c]{90}{Fine}}
         & $\text{TL}_{\text{SH}}$~\cite{wu2017sampling} & 79.8 & 59.6 & 14.5 & 66.3 & 33.3 & 51.5 \\
         & NSM~\cite{norm_softmax} & 81.3 & 61.3 & 13.4  & 70.2 & \underline{37.6} & 38.8 \\
         & NCA++~\cite{proxynca++} & 81.4 & 61.7 & 13.6 & 67.3 & 37.0 & 44.5 \\
         & \textcolor{black}{Smooth-AP~\cite{smoothap}} & 81.3 & 61.7 & 13.4 & 67.3 & 35.2 & 53.1 \\
         & ROADMAP~\cite{ramzi2021robust} & \textbf{82.2} & \textbf{62.5} & 12.9 & 69.3 & 35.1 & 50.4 \\
         
         \midrule
         \multirow{3}{*}{\rotatebox[origin=c]{90}{Hier.}}
         & CSL~\cite{sun2021dynamic} & 79.4 & 58.0 & \underline{45.0} & 62.9 & 30.2 & \underline{88.5} \\
         \cmidrule{2-8}
         & \textbf{HAPPIER} & 81.0 & 60.4 & \textbf{58.4} & \underline{70.7} & 36.7 & \textbf{88.6} \\
         & \textbf{$\text{HAPPIER}_{\text{F}}$} & \underline{81.8} & \underline{62.2} & 36.0 & \textbf{71.6} & \textbf{37.8} & 85.1 \\
        \bottomrule
    \end{tabularx}
\end{table*}

\subsubsection{Relevance function choice}

\begin{table*}[t]
% \vspace{-20pt}
\begin{minipage}[t]{0.49\textwidth}
    % \vspace{0pt}
    % \vspace{-0.8\intextsep}
    % \setlength\tabcolsep{0.3pt}
    % 
    %  
        \caption{Impact of optimization choices for $\hap$ (cf.~\cref{sec:suprank}) on iNat-base.}
     \label{tab:hrank_optim}
    %  \vspace{2pt}
    \centering
    \begin{tabularx}{1\textwidth}{YYY }
        \toprule
         $\lhaps$ & \multirow{1}{*}{$\Ldecom$} & $\hap$ \\
         \midrule
        % \multirow{4}{*}{\rotatebox[origin=c]{90}{\HAPPIER}}
        \xmark & \xmark & 52.3 \\
        \cmark & \xmark & 53.1 \\
        \cmark & \cmark & \textbf{54.3} \\
         \bottomrule
    \end{tabularx}
\end{minipage}%
\hfill
\begin{minipage}[t]{0.49\textwidth}
    % \vspace{0pt}
    % \vspace{-0.8\intextsep}
    % \setlength\tabcolsep{0.3pt}
    % 
    %  
    %\centering
    %\setlength\tabcolsep{0pt}
    \caption{Comparison of $\hap$ (\cref{eq:hierarchy_relevance}) and $\Sigma w\AP$ from~\cref{prop:link_hap_ap}.}
        \label{tab:analysis_relevance}
        \centering
    \begin{tabularx}{1\textwidth}{ l YYY}
        \toprule
         \begin{tabular}{@{}l@{}}
                   test$\rightarrow$\\
                   train$\downarrow$\\
                 \end{tabular} & $\hap$ & $\sum w\AP$ & NDCG \\
         \midrule
         $\hap$ & \textbf{53.1} & 39.8 & \textbf{97.0} \\
        $\sum w\AP$ & 52.0 & \textbf{40.5} & 96.4 \\
         \bottomrule
    \end{tabularx}
\end{minipage}%
% \vspace{-20pt}
\end{table*}

\cref{tab:analysis_relevance} compares models that are trained with the relevance function of~\cref{eq:hierarchy_relevance}, \ie $\hap$, and $\sum w\AP$ (relevance given in~\cref{prop:link_hap_ap}). We report results for $\hap$, $\sum w\AP$ and NDCG. Both $\hap$, $\sum w\AP$ perform better when trained with their own metric: +1.1pt $\hap$ for the model trained to optimize it and +0.7pt $\sum w\AP$ for the model trained to optimize it. Both models show similar performances in NDCG (96.4 \vs 97.0).

\subsubsection{Choices of optimization}

In~\cref{tab:hrank_optim}, we study the impact of our different choices regarding the direct optimization of $\hap$. The baseline method uses a sigmoid to optimize $\hap$ as in~\cite{Qin2009AGA,smoothap}. Switching to our surrogate loss $\lhaps$ \cref{sec:suprank} yields a +0.8pt increase in $\hap$. Finally, the combination with $\lclust$ in HAPPIER results in an additional 1.3pt improvement in $\hap$.

\subsection{Additional qualitativ results}

\begin{figure*}[t]
    \centering
        
    \begin{subfigure}[t]{\textwidth}
        \includegraphics[width=0.9\textwidth]{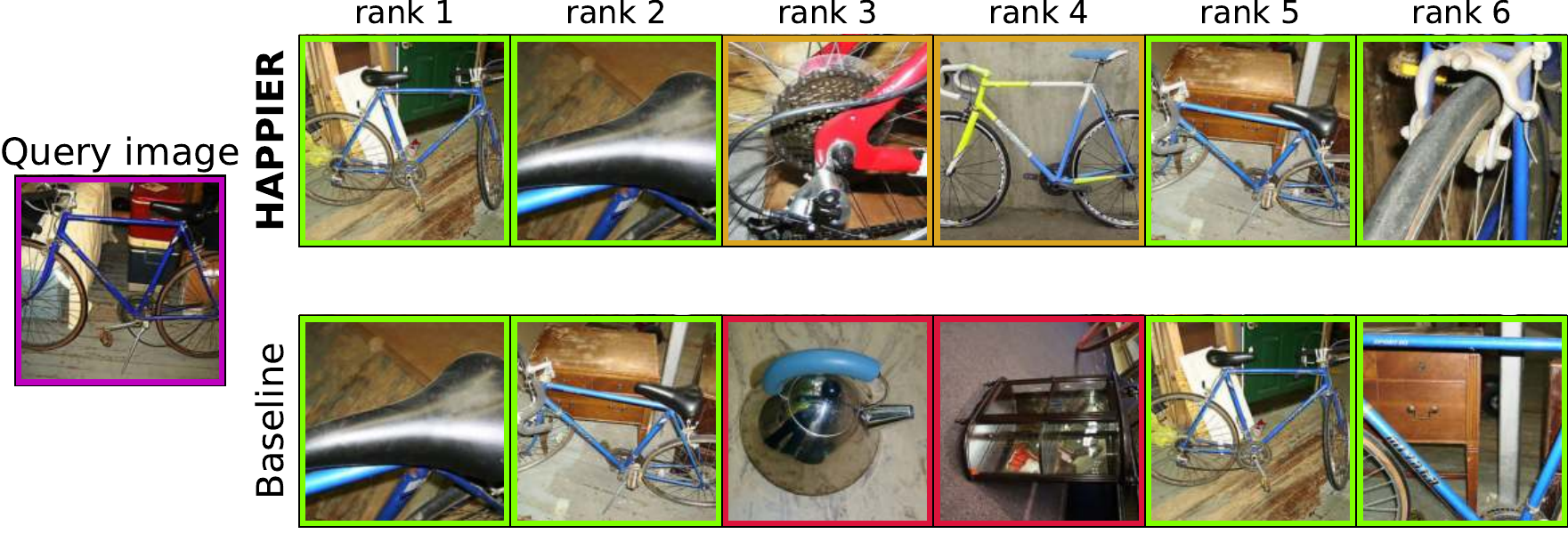}
        \caption{HAPPIER can help make less severe mistakes. The inversion on the bottom row are with negative instances (in \textcolor{red}{red}), where as with HAPPIER (top row) inversions are with instances sharing the same coarse label ``bike'' (in \textcolor{orange}{orange}).}
        \label{fig:qual_sop_good}
    \end{subfigure}
    
    \begin{subfigure}[t]{\textwidth}
    \includegraphics[width=0.9\textwidth]{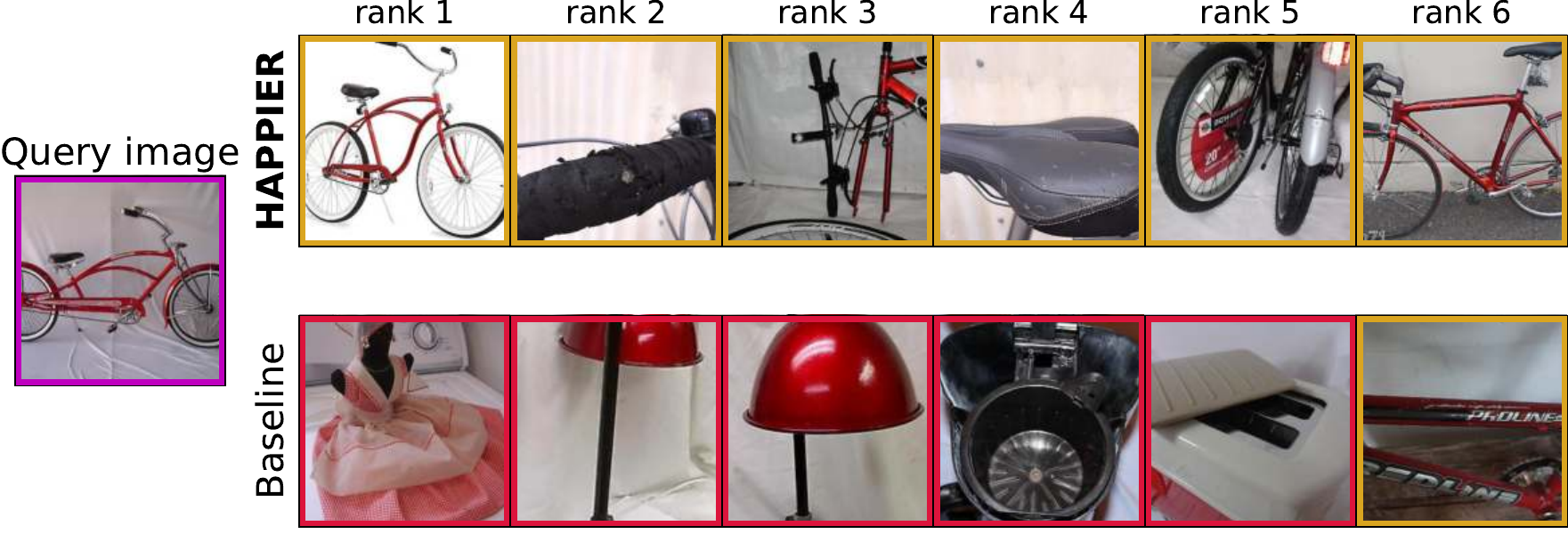}
    \caption{In this example, the models fail to retrieve the correct fine grained images. However HAPPIER still retrieves images of very similar bikes (in \textcolor{orange}{orange}) whereas the baseline retrieves images that are dissimilar semantically to the query (in \textcolor{red}{red}).}
    \label{fig:qual_sop_error}
    \end{subfigure}%

    \caption{Qualitative examples of failure cases from a standard fine-grained model corrected by training with HAPPIER.}
    \label{fig:qualitative_results}
\end{figure*}

\noindent\textbf{Controlled errors: SOP} We showcase in~\cref{fig:qualitative_results} errors of HAPPIER \vs a fine-grained baseline. On~\cref{fig:qual_sop_good}, we illustrate how a model trained with HAPPIER makes mistakes that are less severe than a baseline model trained only on the fine-grained level. On~\cref{fig:qual_sop_error}, we show an example where both models fail to retrieve the correct fine-grained instances, however the model trained with HAPPIER retrieves images of bikes that are visually more similar to the query.

%angle=90,

% \end{IEEEbiography}


% % insert where needed to balance the two columns on the last page with
% %\newpage



%\vfill


\end{document}